\pdfoutput=1
\documentclass[onecolumn]{tech_report/forgewm_arxiv}

\usepackage{fontawesome5}
\usepackage{enumitem}
\usepackage{placeins}
\usepackage{capt-of}
\usepackage{tabularx}

\title{%
\makebox[0pt][l]{%
  \hspace{-2.3em}%
  \smash{\raisebox{-1.35em}{%
    \includegraphics[height=3.6em]{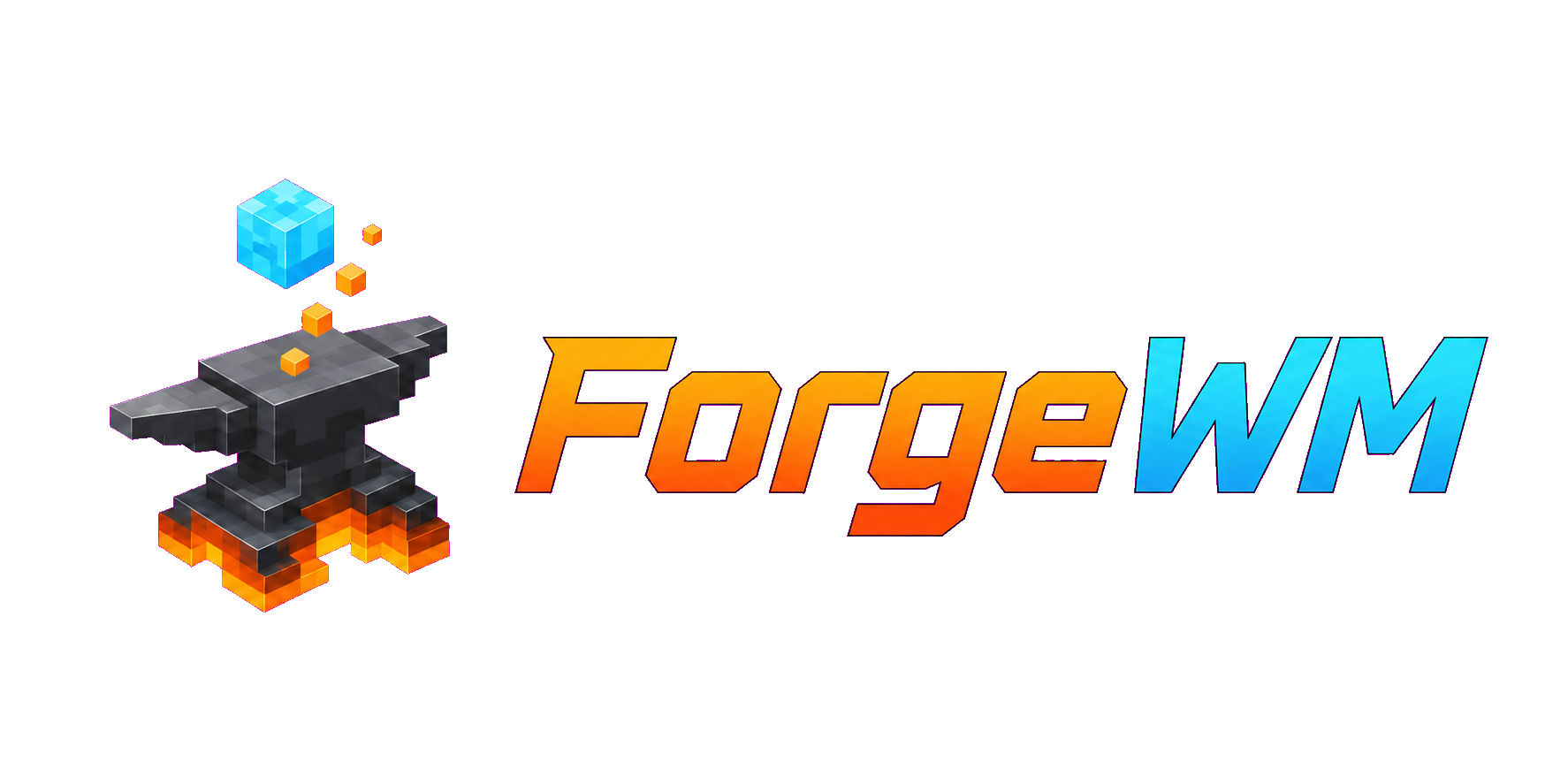}%
  }}%
}%
\phantom{ForgeWM}: Progressive Causal Training for Few-Step Action-Conditioned
Video World Models}

\author[1,2]{Xinye Li$^{*}$}
\author[3,4]{Lingshuai Lin$^{*}$}
\author[2]{Lei Wang}
\author[5]{Liuzhou Zhang}
\author[3,4]{Jialin Cui}
\author[4]{Qingshan Li}
\author[4]{Guanchu Wang}
\author[2]{Qingbin Liu}
\author[2]{Xi Chen}
\author[2]{Jiang Bian}
\author[1]{Wai Lam$^{\dagger}$}

\affiliation[1]{CUHK}
\affiliation[2]{Tencent PCG}
\affiliation[3]{FDU}
\affiliation[4]{Shanghai AI Laboratory}
\affiliation[5]{HKUST}

\contribution[*]{Equal contribution}
\contribution[\dagger]{Corresponding author}
\contribution[]{\faEnvelope\ Contact: Xinye Li (\email{xyli@se.cuhk.edu.hk})}

\reportshorttitle{ForgeWM: Progressive Causal Training for Few-Step World Models}

\metadata[Resources]{%
  \href{https://asdfo123.github.io/ForgeWM}{\faGlobe~Page}\hspace{1.6em}%
  \href{https://github.com/asdfo123/ForgeWM}{\faGithub~Code}\hspace{1.6em}%
  \href{https://huggingface.co/ForgeWM}{\faCube~Models}%
}

\abstract{
Action-conditioned video world models require low-latency causal generation and reliable responses to game-native controls. Although causal distillation enables one- or few-step video synthesis, extending it to interactive world models remains challenging, as discrete keyboard states and continuous mouse motion must remain aligned with temporally compressed latent chunks during causal training and autoregressive rollout. We introduce \emph{ForgeWM}, a progressive framework that transforms a bidirectional action-conditioned video generator into efficient few-step world models through domain adaptation, teacher-forced causal training, causal consistency distillation, and on-policy distribution matching with a bidirectional teacher. The resulting budget-specialized students operate at steady-state denoising budgets of 1, 2, and 4 steps. ForgeWM further supports a dual-path deployment protocol combining latency-critical interaction with optional replay-time refinement, where the one-step student re-noises and refines its saved draft. On paired Minecraft trajectories, ForgeWM leads the evaluated systems in Imaging Quality, reference-aligned motion-profile agreement, action-sign accuracy, and mouse-control accuracy, while achieving the lowest reference LPIPS; the same four-stage recipe transfers to gamepad-controlled FPS gameplay. Replay-time refinement matches four-step reference quality while remaining roughly three times closer to the experienced trajectory than regeneration from noise. These results demonstrate ForgeWM's effectiveness for controllable few-step video generation.

}

\aftertitle{%
  
  {%
    \centering
    \begin{minipage}{0.94\textwidth}
      \centering
      \includegraphics[width=\linewidth]{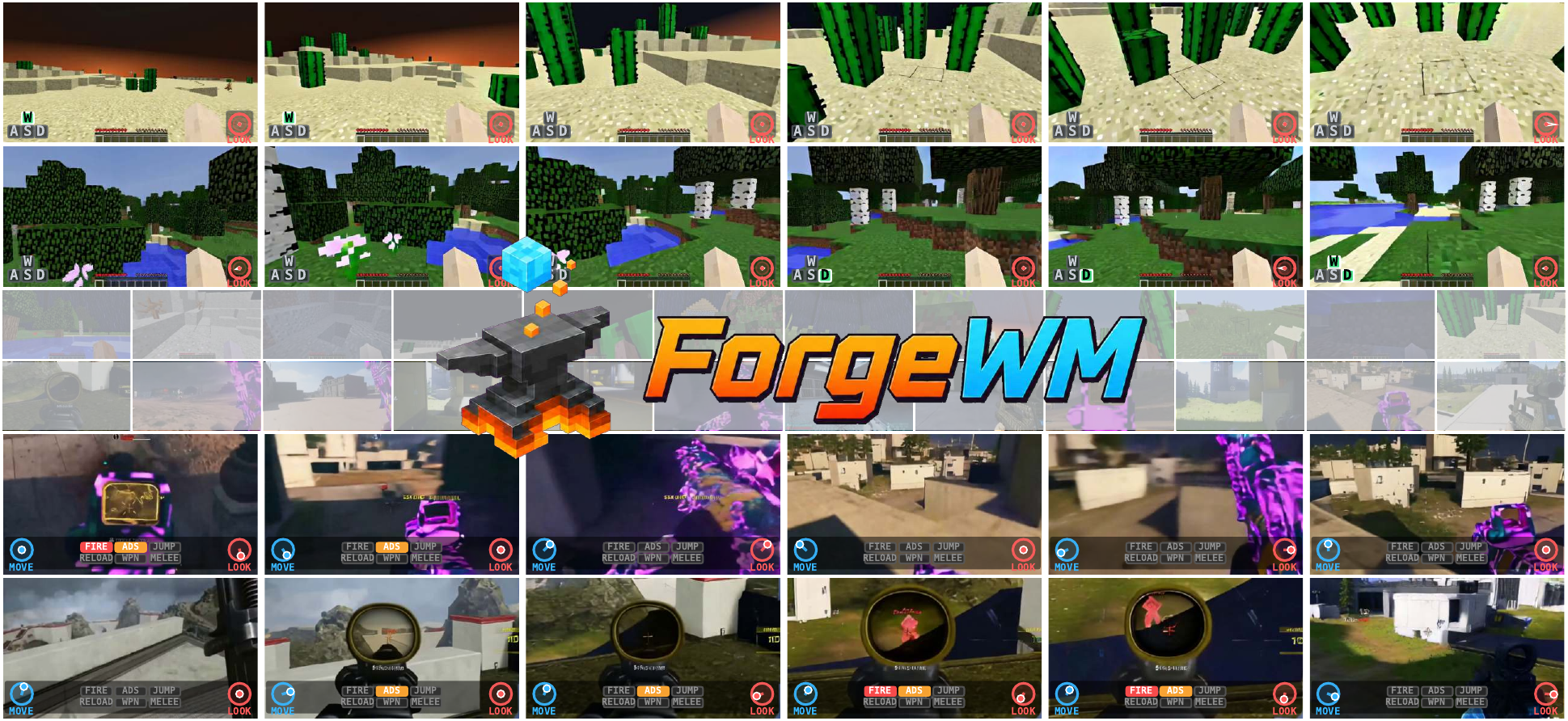}
    
      \captionof{figure}{\textbf{ForgeWM rollouts.} Minecraft (top) under
      keyboard--mouse control and an FPS domain (bottom) under gamepad control,
      both produced by budget-specialized few-step students trained with the same
      four-stage framework.}
      \label{fig:teaser}
    \end{minipage}
    \par
  }%
}

\begin{document}

\maketitle

\section{Introduction}
\label{sec:introduction}

Generative video models can simulate future visual states from observations
and actions without an explicit graphics engine. Interactive deployment,
however, requires causal generation, persistent action responsiveness, and
few enough denoising steps to close the control loop. Recent systems
demonstrate autoregressive generation with keyboard, mouse, or camera-pose
control~\cite{he2025matrixgame2,wang2026matrixgame3,
gao2026lingbotworld2,dreamx2026world}. Yet aggressive sampling compression feeds
visual, action, and cache errors into subsequent chunks, coupling fidelity,
controllability, and multi-chunk stability.
Figure~\ref{fig:teaser} shows ForgeWM rollouts in two control domains.

The difficulty arises because causal generation changes both the input
distribution and model state. At inference, the denoiser conditions on
imperfect self-generated visual history rather than clean data, while action
histories and key--value caches must remain synchronized with the generated
latent chunks. Fewer denoising steps amplify errors in these states, which
then propagate autoregressively. This coupling is especially pronounced for
game-native controls: discrete keyboard states and continuous mouse motion
arrive at video-frame rate and enter the denoiser through dedicated pathways
rather than solely as a camera trajectory. A usable training framework must
therefore preserve the control interface, action-to-latent alignment, and
causal state-update protocol throughout adaptation and distillation.

Recent causal video methods~\cite{huang2025selfforcing,zhao2026causalforcingpp,
zheng2026causalrcm} combine teacher-forced causalization, few-step
distillation, and on-policy self-rollout. ForgeWM studies how to preserve
frame-rate discrete and continuous game controls through this conversion as
clean context gives way to generated history.

To address this challenge, we introduce \emph{ForgeWM}, a progressive
four-stage framework for few-step action-conditioned video world models. It
converts a bidirectional generator through domain adaptation, teacher-forced
causal training, causal consistency distillation, and on-policy distribution
matching. The bidirectional teacher and causal student share a base
initialization and meet during autoregressive self-rollout, where the former
supervises the consistency-initialized student. A modular action interface
preserves aligned keyboard-and-mouse conditioning throughout training. We
train budget-specialized 1-, 2-, and 4-step students, providing distinct
quality--latency operating points.

We evaluate ForgeWM against interactive world-model baselines at its native
1-, 2-, and 4-step operating points, and isolate test-time solver scaling
using a frozen one-step checkpoint. We further compare draft-preserving replay
with from-noise generation, separating online interaction from optional
replay-time refinement.

The contributions of this work are:
\begin{itemize}[leftmargin=1.2em,itemsep=1pt,topsep=3pt,parsep=0pt]
    \item We introduce a four-stage framework that converts a bidirectional
    action-conditioned generator into budget-specialized few-step causal
    world models.

    \item We preserve frame-aligned discrete and continuous game-native controls
    across latent compression, causal training, and autoregressive rollout.

    \item We provide 1-, 2-, and 4-step operating points and a
    same-checkpoint replay protocol that improves offline quality while
    retaining the experienced trajectory.
\end{itemize}

\section{Related Work}
\label{sec:related-work}

\paragraph{Interactive video world models.}
Early diffusion-based simulators such as DIAMOND and GameNGen model game
observations conditioned on agent actions
\cite{alonso2024diamond,valevski2024gamengen}, while Genie learns
action-controllable environments from unlabeled video~\cite{bruce2024genie}
and MineWorld targets real-time Minecraft interaction
\cite{guo2025mineworld}. Recent systems extend this setting with frame-level
controls, long-horizon memory, and cross-game transfer
\cite{he2025matrixgame2,wang2026matrixgame3,tong2026scope}.
LingBot-World 2.0 combines MoBA causal pretraining with consistency and
rollout-level distribution matching~\cite{gao2026lingbotworld2}, while
Cosmos 3 develops an omnimodal world-model backbone for Physical AI
\cite{nvidia2026cosmos3}.

A complementary family parameterizes interaction through camera geometry.
PRoPE encodes relative projective transformations in attention
~\cite{li2025prope}, minWM combines camera-controllable adaptation with causal
distillation~\cite{zhao2026minwm}, and WorldPlay uses dual action
representations with reconstituted context memory~\cite{worldplay2025}.
DreamX-World 1.0 extends this line with E-PRoPE, causal forcing, long-rollout
distribution matching, and geometry-retrieved scene memory
~\cite{dreamx2026world}.  Relative to these camera-, prompt-, or omnimodal
systems, ForgeWM centers frame-aligned keyboard-and-mouse controls across
budget-specialized causal students and draft-preserving replay.

\paragraph{Few-step autoregressive video distillation.}
One- and few-step generation builds on consistency models, which learn direct
noise-to-data mappings~\cite{song2023consistency}, and distribution matching
distillation, which aligns a fast student's distribution with that of a
diffusion teacher~\cite{yin2024dmd}. Recent video methods adapt these ideas to causal autoregressive generation. Diffusion Forcing assigns independent
noise levels to sequence tokens~\cite{chen2024diffusionforcing}, while
Self-Forcing reduces exposure bias through autoregressive self-rollout
\cite{huang2025selfforcing} and AAPT provides an adversarial student-forcing
alternative for one-NFE generation~\cite{lin2025aapt}.

Causal Forcing initializes the student along a causal teacher trajectory
before asymmetric distribution matching~\cite{zhu2026causalforcing}.
Causal Forcing++ replaces offline ODE pairs with online causal consistency
distillation~\cite{zhao2026causalforcingpp}, and Causal-rCM unifies
teacher-forcing consistency learning with Self-Forcing distribution matching
\cite{zheng2026causalrcm}. ForgeWM instantiates this progression in a
game-native training-to-deployment framework, producing separate 1-, 2-, and
4-step causal checkpoints with aligned discrete and continuous controls.

Most fixed-budget models can be evaluated with additional solver steps, but
quality need not improve outside their trained regime. AnyFlow instead learns
flow-map transitions for flexible test-time budgets~\cite{gu2026anyflow};
ForgeWM trains budget-specialized students and separately evaluates a frozen
one-step checkpoint under off-budget schedules.

\paragraph{Replay-time and video refinement.}
Adding noise to an existing sample and denoising it with a generative prior is
the basis of SDEdit~\cite{meng2021sdedit}.  Recent video systems develop more
specialized correction mechanisms.  AutoRefiner learns pathwise noise
refinement and a reflective cache for autoregressive video diffusion
~\cite{yu2025autorefiner}; Pathwise Test-Time Correction uses the initial
frame to calibrate intermediate stochastic states without retraining
~\cite{xiang2026ttc}; and SANA-WM employs a trained long-video refiner in a
two-stage minute-scale pipeline~\cite{zhu2026sanawm}. ForgeWM reconditions a saved rollout with the deployed low-step student at
a larger schedule, requiring no dedicated refiner or second checkpoint.  We evaluate reference quality and draft retention separately.

\begin{figure}[t]
    \centering
    \includegraphics[width=\textwidth]{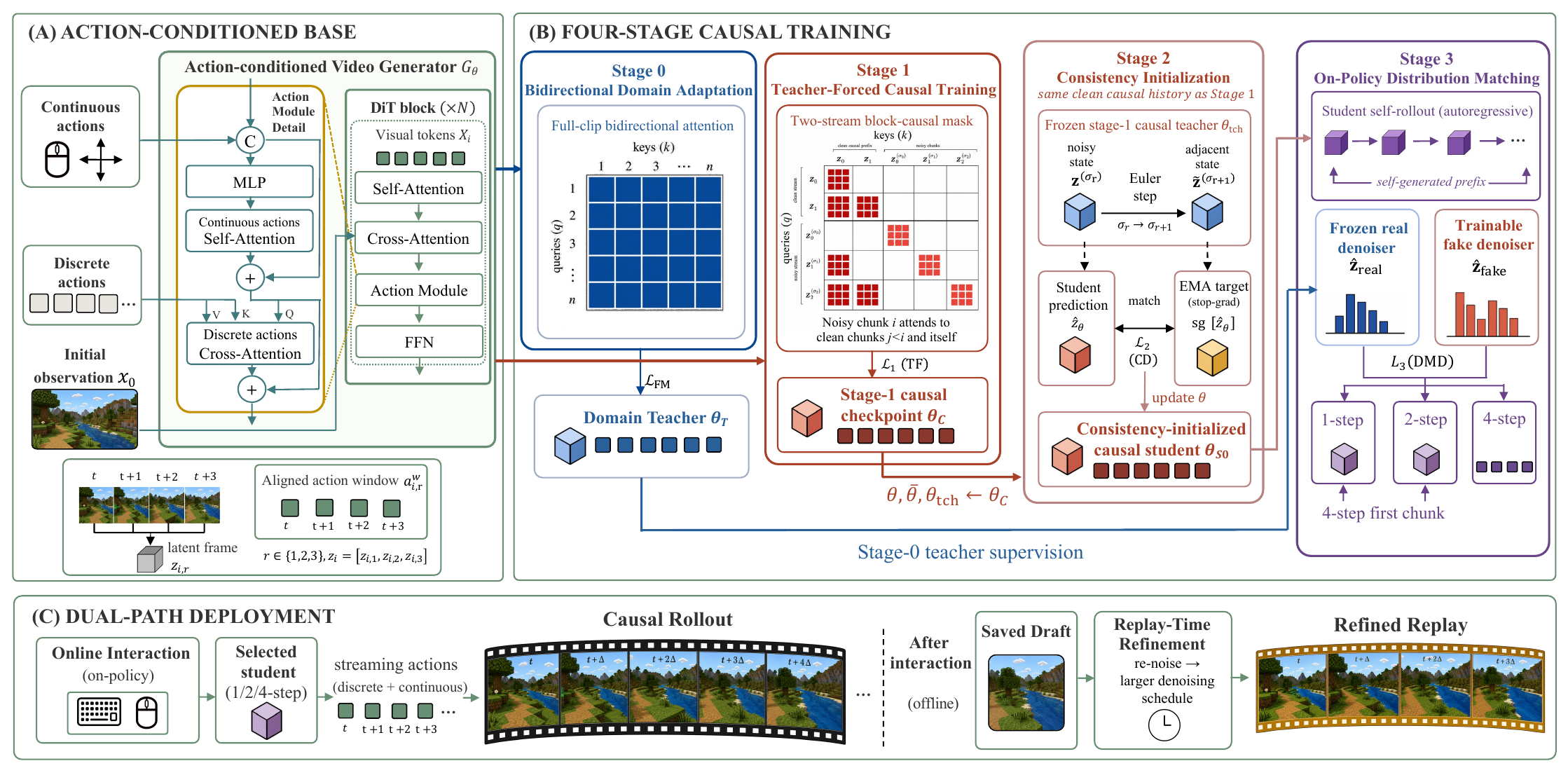}
    \caption{\textbf{Overview of ForgeWM.}  (A) Frame-aligned keyboard and
    mouse controls condition latent chunks.  (B) A shared base yields a
    bidirectional teacher and budget-specialized causal students through
    four-stage training.  (C) Deployment separates low-latency interaction
    from optional Replay-Time Refinement (Figure~\ref{fig:replay-refinement}).}
    \label{fig:overview}
\end{figure}
\section{Method}
\label{sec:method}

ForgeWM transforms a bidirectional action-conditioned video generator into
causal world models specialized for different sampling budgets.
Figure~\ref{fig:overview} summarizes its game-native action interface,
four-stage training graph, and dual-path deployment.  Two branches start from
the same base generator: a bidirectional domain teacher and a causal student.
Their outputs meet during on-policy distribution matching over the
student's autoregressive self-rollouts.  We next describe
the shared generator and action interface, followed by the training and
deployment procedures.

\subsection{Problem Setup and Base Generator}

Let $x_{0:L}$ be a video, $x_0$ its initial observation, and
$a_{0:L}=(k_{0:L},m_{0:L})$ the corresponding control stream, where $k$ is a
discrete keyboard state and $m=(\Delta u,\Delta v)$ is continuous mouse
motion.  A fixed VAE maps the video to a clean latent sequence $z$.  Following
the Matrix-Game 2.0 image-to-video design~\cite{he2025matrixgame2}, the
generator receives the encoded first frame through both channel-wise latent
conditioning and a visual-context branch.

We adapt the base generator using the flow-matching objective
~\cite{lipman2022flow}.  Using the noise level $\sigma\in[0,1]$ as the path
coordinate and Gaussian noise $\epsilon$, we construct
\begin{equation}
    z_\sigma=(1-\sigma)z+\sigma\epsilon,
    \qquad v^*(z_\sigma,\sigma)=\epsilon-z,
\end{equation}
and train the velocity predictor $v_\theta$ using
\begin{equation}
    \mathcal{L}_{\mathrm{FM}}
    =\mathrm{E}_{z,\epsilon,\sigma}\!
    \left[w(\sigma)\left\|v_\theta(z_\sigma,\sigma;c)
    -v^*(z_\sigma,\sigma)\right\|_2^2\right],
\end{equation}
where $c$ contains the initial-frame and action conditions.  The induced clean
prediction is
\begin{equation}
    \hat z_\theta\!\left(z_\sigma,\sigma\right)
    =z_\sigma-\sigma\,v_\theta\!\left(z_\sigma,\sigma\right),
    \label{eq:clean-pred}
\end{equation}
which is the quantity the consistency and distribution-matching stages compare.
The first stage
uses bidirectional temporal attention over each training clip; later stages
change the temporal execution and training distribution while retaining this
conditioning interface.

\subsection{Game-Native Action Conditioning}

Following Matrix-Game 2.0~\cite{he2025matrixgame2}, the
\textit{ActionModule} injects frame-rate discrete actions and continuous
controls through separate pathways, without reducing them to camera poses or
pose-based encodings such as PRoPE~\cite{li2025prope}. Discrete inputs serve as
cross-attention keys and values, while windowed continuous inputs are fused
with visual features and processed by temporal attention.

The VAE compresses four video frames into one latent frame, and each
three-latent causal chunk therefore spans twelve video frames. We group actions
by latent interval and retain the preceding controls required by each token's
temporal window. During rollout, the model maintains a visual key--value cache
together with separate keyboard and mouse caches at action-enabled blocks, so
each new chunk receives its complete aligned action window. The same alignment
and cache-update protocol is used throughout causal training, distillation,
and inference.

\subsection{Four-Stage Causal Training}
\label{sec:four-stage}

ForgeWM progressively changes the temporal execution pattern, sampling
objective, and conditioning-history distribution across four stages.
\begin{table}[h]
    \centering
    \small
    \setlength{\tabcolsep}{2.2pt}
    \begin{tabular*}{\textwidth}{@{\extracolsep{\fill}}cllll@{}}
        \toprule
        \textbf{Stage} & \textbf{Initialization}
        & \textbf{Training Context}
        & \textbf{Objective} & \textbf{Output} \\
        \midrule
        0 & Base    & Full-clip bidirectional & FM        & Domain teacher \\
        1 & Base    & Clean causal history    & Causal FM & Causal teacher \\
        2 & Stage 1 & Clean causal history    & Online CD & Few-step initializer \\
        3 & Stage 2 & Self-generated history  & DMD       & 1/2/4-step students \\
        \bottomrule
    \end{tabular*}
    \caption{\textbf{Training stages.}
    Each stage changes either the temporal execution pattern,
    sampling objective, or history distribution.}
    \label{tab:training-stages}
\end{table}

Table~\ref{tab:training-stages} provides a roadmap of the four stages;
Figure~\ref{fig:overview}(B) visualizes their computational dependencies,
and Appendix~\ref{app:training-details} reports the corresponding
optimization settings and stage-wise ablations.

\paragraph{Stage 0: bidirectional adaptation.}
We first adapt the action-conditioned image-to-video generator to the target
game domain using $\mathcal{L}_{\mathrm{FM}}$.  Full-clip bidirectional
attention learns the visual and control prior.  We retain this
checkpoint as the frozen real denoiser used during Stage~3.

\paragraph{Stage 1: teacher-forced causal training.}
In parallel to Stage~0, we initialize a second branch from the same base
generator and replace full temporal attention with block-wise causal
attention.  Frames inside a latent chunk attend bidirectionally, while each
chunk can attend only to earlier chunks.  Training concatenates the clean and
noisy token streams and applies a block mask under which noisy chunk $i$
attends to clean chunks $j<i$ and to itself, so the history is exact rather
than model-generated.  A single noise level is drawn per chunk, giving the
teacher-forced objective
\begin{equation}
    \mathcal L_{1}
    =\mathrm{E}\!\left[w(\sigma_i)\left\|
    v_\theta\!\left(z^{(\sigma_i)}_i,\sigma_i;c_i,z_{<i}\right)
    -(\epsilon_i-z_i)\right\|_2^2\right],
    \label{eq:stage1}
\end{equation}
where $z_{<i}$ are the clean preceding chunks, $c_i$ is the initial-frame and
action condition for chunk $i$, and $w$ is the same flow-matching weighting as
Stage~0.  This stage learns the causal execution pattern without yet exposing
the model to errors from its own rollout.

\paragraph{Stage 2: causal consistency initialization.}
Starting from the Stage~1 causal checkpoint, we perform online causal
consistency distillation, building on consistency
distillation~\cite{song2023consistency} and its causal video adaptations
~\cite{zhao2026causalforcingpp}.  Generator, its
exponential-moving-average copy, and a frozen teacher are all initialized from
that checkpoint.  We use a discrete grid of $N\!=\!48$ noise levels
$\sigma_0>\dots>\sigma_{N-1}$ and sample one adjacent pair $(\sigma_i,
\sigma_{i+1})$ per step.  The frozen teacher takes one classifier-free-guided
Euler step from $\sigma_i$ towards $\sigma_{i+1}$,
\begin{equation}
    \tilde z^{(\sigma_{i+1})}
    =z^{(\sigma_i)}
    +\left(\sigma_{i+1}-\sigma_i\right)
    \left[v^{\emptyset}_{\mathrm{tch}}
    +\omega\left(v^{c}_{\mathrm{tch}}-v^{\emptyset}_{\mathrm{tch}}\right)
    \right],
    \label{eq:stage2-euler}
\end{equation}
with guidance weight $\omega$ and $v^{c}_{\mathrm{tch}}$, $v^{\emptyset}_{\mathrm{tch}}$
the conditional and unconditional teacher velocities.  Because
$\sigma_{i+1}<\sigma_i$, the step moves towards the data end of the path.  The
student's original-level clean prediction is then matched to the EMA
stop-gradient at the teacher-advanced point:
\begin{equation}
    \mathcal L_{2}
    =\mathrm{E}\!\left[\left\|
    \hat z_\theta\!\left(z^{(\sigma_i)},\sigma_i\right)
    -\mathrm{sg}\!\left[
    \hat z_{\bar\theta}\!\left(\tilde z^{(\sigma_{i+1})},\sigma_{i+1}\right)
    \right]\right\|_2^2\right],
    \label{eq:stage2}
\end{equation}
where $\bar\theta$ is the EMA parameter vector and $\mathrm{sg}[\cdot]$ denotes
stop-gradient.  All three networks are conditioned on clean causal history at
this stage, which is what lets the teacher's Euler step be evaluated in one
forward pass instead of an autoregressive unroll.  This local consistency
objective initializes few-step sampling without an offline trajectory dataset.

\paragraph{Stage 3: on-policy distribution matching.}
Finally, the student performs autoregressive self-rollout, so each new causal
chunk is conditioned on its own previously generated history.  The student is
initialized from Stage~2, while the domain-adapted Stage~0 checkpoint provides
the real-distribution supervision.  Let $\hat z$ be the clean latent
produced by a $K$-step self-rollout and let $\hat z^{(\sigma)}$ be a re-noised
copy of it.  Following distribution matching distillation~\cite{yin2024dmd} and its
autoregressive video adaptations in Self-Forcing and
Causal Forcing~\cite{huang2025selfforcing,zhu2026causalforcing}, we form the
gradient direction from the disagreement between a frozen real denoiser
$\hat z_{\mathrm{real}}$ and a trainable fake denoiser
$\hat z_{\mathrm{fake}}$, which provide score-equivalent predictions under the
denoiser parameterization of Eq.~(\ref{eq:clean-pred}),
\begin{equation}
    g=\frac{\hat z_{\mathrm{fake}}\!\left(\hat z^{(\sigma)},\sigma\right)
    -\hat z_{\mathrm{real}}\!\left(\hat z^{(\sigma)},\sigma\right)}
    {\mathrm{mean}\!\left(\left|\hat z-\hat z_{\mathrm{real}}
    \!\left(\hat z^{(\sigma)},\sigma\right)\right|\right)},
    \label{eq:stage3-grad}
\end{equation}
where the denominator averages the absolute deviation over all elements of each
sample and acts as an adaptive normalizer.  We apply this direction through the
generator surrogate
\begin{equation}
    \mathcal L_{3}
    =\mathrm{E}\!\left[\frac{1}{2}
    \left\|\hat z-\mathrm{sg}\!\left[\hat z-g\right]\right\|_2^2\right],
    \label{eq:stage3}
\end{equation}
whose gradient with respect to $\hat z$ is exactly $g$; the fake denoiser is
updated concurrently with its own flow-matching loss on the student's samples.
Training on the induced rollout
distribution reduces the clean-history mismatch left by teacher forcing.  We
train separate budget-specialized students for
$K\in\{1,2,4\}$.  Following the
first-frame enhancement strategy of~\citet{yang2025asd}, the 1- and 2-step
students use a fixed four-step schedule for the first generated latent chunk
and the matched $K$-step schedule thereafter; the 4-step student uses four
denoising evaluations throughout.  Thus, ``1-step''
and ``2-step'' denote the steady-state denoising budget rather than the total
number of generator calls for the complete rollout.

\subsection{Budget-Specialized Interaction}

For every generated chunk, the model follows its configured denoising schedule
and appends the clean prediction to the causal history.  The primary deployment
uses the student matched to each target budget.  We distinguish this native
evaluation from an off-budget analysis that freezes the one-step student and
changes only its test-time solver schedule.  Because the low-step models retain
a four-evaluation first chunk, ``one-step'' and ``two-step'' refer to the
steady-state budget.  The fixed-checkpoint step-scaling analysis reports both
generator evaluations and wall time under one consistent inference path.

\paragraph{Dual-path deployment.}
ForgeWM uses the native low-step student for online interaction and optionally
applies a larger denoising budget to the saved rollout after interaction. The
replay path starts from the realized draft rather than from fresh noise and
reuses the recorded actions and refined causal prefix. The empirical motivation, formulation, and evaluation of this replay path are
presented together later under \emph{Inference Scaling and Replay-Time
Refinement} (Section~\ref{sec:scaling-replay}).

\section{Experiments}
\label{sec:experiments}

\subsection{Implementation Details}

\paragraph{Training.}
We initialize ForgeWM from the public Matrix-Game 2.0
lineage~\cite{he2025matrixgame2} and train on 40,000 clips constructed from
GF-Minecraft~\cite{yu2025gamefactory} at $640\!\times\!352$. The final stage
produces students with steady-state budgets of one, two, and four denoising
updates. We also compare checkpoints from Stages 1--3 under a shared four-step
inference schedule.
Full optimization hyperparameters, stage-wise training details, and ablation
results are provided in Appendix~\ref{app:training-details}.

\paragraph{Evaluation.}
We compare ForgeWM with Matrix-Game 2.0~\cite{he2025matrixgame2} and the
HY-WorldPlay checkpoint of WorldPlay~\cite{worldplay2025} using 77-frame
rollouts with shared initial frames and controls. Both baselines use four
denoising updates per chunk; ForgeWM uses four updates for the initial chunk
and its native one-, two-, or four-step budget thereafter. ForgeWM and
Matrix-Game 2.0 share a keyboard-and-mouse interface, while a deterministic
adapter maps the same controls to HY-WorldPlay's parameterization.
Reference-aligned metrics use 1{,}000 paired trajectories; no-reference metrics
use 462 constant-action videos from 77 initial states. Chunk time is the mean
sampling time over non-initial chunks, excluding loading, VAE decoding, and
writing; FPS accounts for the 12-frame ForgeWM/Matrix-Game chunks and 16-frame
HY-WorldPlay chunks. Replay uses frozen ForgeWM-1 at $r=0.3$ with four updates; the direct
four-step baseline uses native ForgeWM-4 generation from noise. The user study includes 41
participants and 615 selections; CrossFPS uses 25 clips from each of seven
games. Full protocols are provided in Appendix~\ref{app:eval-protocol}.

\paragraph{Metrics.}
We report VBench Imaging Quality (IQ), Aesthetic Quality (AQ), and Subject
Consistency~\cite{huang2023vbench}; LPIPS~\cite{zhang2018lpips};
optical-flow profile similarity; and Mouse Accuracy based on GameWorld
Score~\cite{zhang2025matrixgame}. For keyboard control, we report KCtrl, a
camera-trajectory sign test on opposite-action pairs, in place of GameWorld
Score Keyboard Accuracy, whose real-frame-trained inverse-dynamics evaluator
can conflate control response with visual-domain similarity. For replay,
Replay LPIPS and $D_{\mathrm{draft}}$ use LPIPS with an AlexNet backbone on
the same 16 evaluated frames, excluding the initial frame. Replay LPIPS
compares the output with its paired reference, whereas
$D_{\mathrm{draft}}$ compares it with the saved draft. A blind three-way
study measures visual, action, and spatiotemporal preferences.

\subsection{Interactive World Model Comparison}
\begin{figure}[t]
    \centering
    \includegraphics[width=\textwidth]{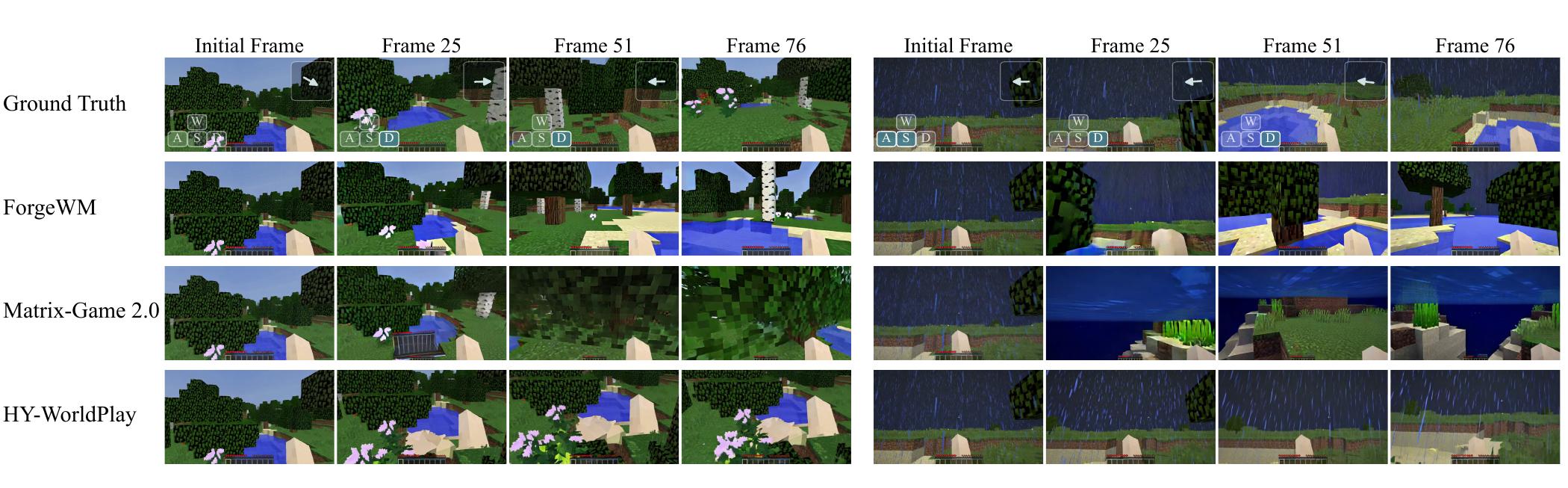}
    \caption{\textbf{Qualitative comparison.} Rollouts at frames 0, 25, 51,
    and 76.  Rows show the reference and three models.  Controls are
    annotated once on the reference: each overlay shows dominant WASD and
    accumulated mouse-look to the next frame; the final frame has no
    outgoing control overlay.  Left/right: daytime forest stream/rainy
    riverbank at night.}
    \label{fig:qualitative-comparison}
\end{figure}
\begin{table}[t]
    \centering
    \footnotesize
    \setlength{\tabcolsep}{0.6pt}
    \begin{tabular*}{\textwidth}{@{\hspace{0.003\textwidth}\extracolsep{\fill}}l ccccccccc@{\hspace{0.003\textwidth}}}
        \toprule
        \multirow{2}{*}[-0.5ex]{Model}
            & \multicolumn{3}{c}{Visual Quality}
            & \multicolumn{2}{c}{Temporal Quality}
            & \multicolumn{2}{c}{Action Controllability}
            & \multicolumn{2}{c}{Efficiency} \\
        \cmidrule(l{1pt}r{1pt}){2-4}
        \cmidrule(l{1pt}r{1pt}){5-6}
        \cmidrule(l{1pt}r{1pt}){7-8}
        \cmidrule(l{1pt}r{1pt}){9-10}
            & IQ$\uparrow$ & LPIPS$\downarrow$
            & AQ$\uparrow$
            & Subj. Cons.$\uparrow$ & Flow Prof.$\uparrow$
            & KCtrl$\uparrow$ & Mouse Acc.$\uparrow$
            & Latency (ms)$\downarrow$ & FPS$\uparrow$ \\
        \midrule
        Matrix-Game 2.0
            & 0.6282 & 0.6443 & 0.4583 & 0.7349 & 0.9343
            & 0.9156 & 0.7061
            & 370.9 & 32.35 \\
        HY-WorldPlay
            & 0.6133 & 0.6172 & 0.4855
            & \textbf{0.9466} & 0.8288
            & 0.9286 & 0.5818
            & 2164.3 & 7.54 \\
        \midrule
        ForgeWM-1 (1-step)
            & 0.6776 & 0.6529 & 0.4807
            & 0.8279 & 0.9403
            & 0.9545 & 0.7848
            & 168.2 & 72.10 \\
        ForgeWM-2 (2-step)
            & \textbf{0.6865} & 0.6171 & 0.4814
            & 0.8349 & \textbf{0.9429}
            & \textbf{0.9740} & \textbf{0.8268}
            & 239.7 & 50.31 \\
        ForgeWM-4 (4-step)
            & 0.6788 & \textbf{0.6168} & \textbf{0.4860}
            & 0.7613 & 0.9420
            & \textbf{0.9740} & 0.8102
            & 369.6 & 32.47 \\
        \bottomrule
    \end{tabular*}
    \caption{\textbf{Comparison with interactive world models.} Bold marks the
    best quality/control value in each column; efficiency entries are not
    bolded.}
    \label{tab:world-model-comparison}
\end{table}

Table~\ref{tab:world-model-comparison} shows that ForgeWM variants attain
the best reported values in six of seven quality/control columns, while
ForgeWM-1 achieves the highest measured generation throughput.  Performance
is not monotone in the denoising budget, motivating the fixed-checkpoint study
below.  Subject Consistency should be interpreted cautiously because it can
favor conservative, low-motion videos; the reference-aligned Flow Profile
metric better reflects whether the requested motion pattern is reproduced.

ForgeWM-2 and ForgeWM-4 jointly rank first on KCtrl, indicating the highest action-sign accuracy under counterfactual
opposite-action pairs.
Figure~\ref{fig:qualitative-comparison} shows aligned rollouts from the
reference and the three models under a shared initial frame and control trace.
\begin{figure}[!h]
    \centering

    \begin{minipage}[t]{0.48\textwidth}
        \vspace{0pt}
        \emergencystretch=1em

        \paragraph{Human preference study.}
        Figure~\ref{fig:user-study} reports a blind three-way study
        comparing the four-step ForgeWM student with Matrix-Game 2.0
        and HY-WorldPlay. In each comparison, the three clips shared
        the same initial state and control trace; model identities were
        hidden and their left-to-right order was randomized.
        
        Each of the 41 participants evaluated five matched triplets per
        criterion and selected one clip without a tie option, yielding
        205 judgments per criterion and 615 in total. ForgeWM-4 receives
        68.8\% of visual-quality preferences, 57.6\% of action-accuracy
        preferences, and 55.6\% of spatiotemporal-consistency preferences.
        Its pooled preference share is 60.7\%, with the largest margin in
        visual quality. Appendix~\ref{app:user-study} provides recruitment,
interface, and instruction details.
    \end{minipage}
    \hfill
    \begin{minipage}[t]{0.5\textwidth}
        \vspace{0pt}
        \centering
        \includegraphics[width=\linewidth]
            {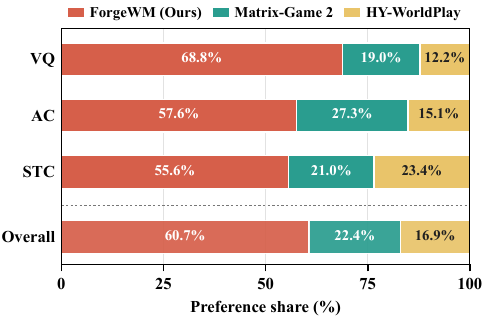}
        \caption{\textbf{Human preferences.}}
        \label{fig:user-study}
    \end{minipage}

\end{figure}

\subsection{Inference Scaling and Replay-Time Refinement}
\label{sec:scaling-replay}

\begin{figure}[!h]
    \centering

    \begin{minipage}[b]{0.48\textwidth}
        
        \emergencystretch=1.5em

        \paragraph{Test-time step scaling.}
        To separate inference-time compute from checkpoint
        specialization, we freeze ForgeWM-1 and vary only its
        denoising schedule. Figure~\ref{fig:nfe-scaling} shows that
        performance remains stable beyond the native one-step budget.

        Imaging Quality peaks at two steps, Subject Consistency at
        four, and reference LPIPS improves through four to eight
        steps before slightly regressing. Flow Profile and KCtrl
        remain stable. More steps primarily increase motion
        magnitude, latency, and computational cost rather than
        directional control.

        \paragraph{From scaling to refinement.}
        These trends reveal a deployment asymmetry: one step is
        preferable for online interaction, whereas additional
        denoising can improve reference-aligned quality without
        sacrificing directional control. Rerunning from noise,
        however, produces a new trajectory rather than improving
        the rollout the user experienced. Replay-Time Refinement
        instead applies the additional compute directly to the
        saved draft.
    \end{minipage}
    \hfill
    \begin{minipage}[b]{0.49\textwidth}
        \vspace{0pt}
        \centering
        \includegraphics[width=\linewidth]{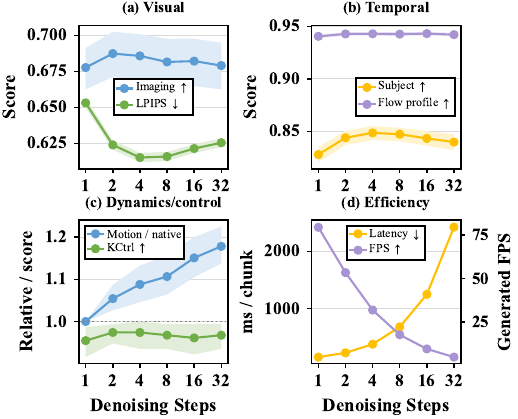}
        \caption{\textbf{Test-time step scaling.}
        ForgeWM-1 under different denoising budgets; shading denotes
        bootstrap 95\% confidence intervals.}
        \label{fig:nfe-scaling}
    \end{minipage}

\end{figure}

\paragraph{Draft-preserving replay.}
\begin{figure}[!ht]
    \centering
    \includegraphics[width=0.93\textwidth]{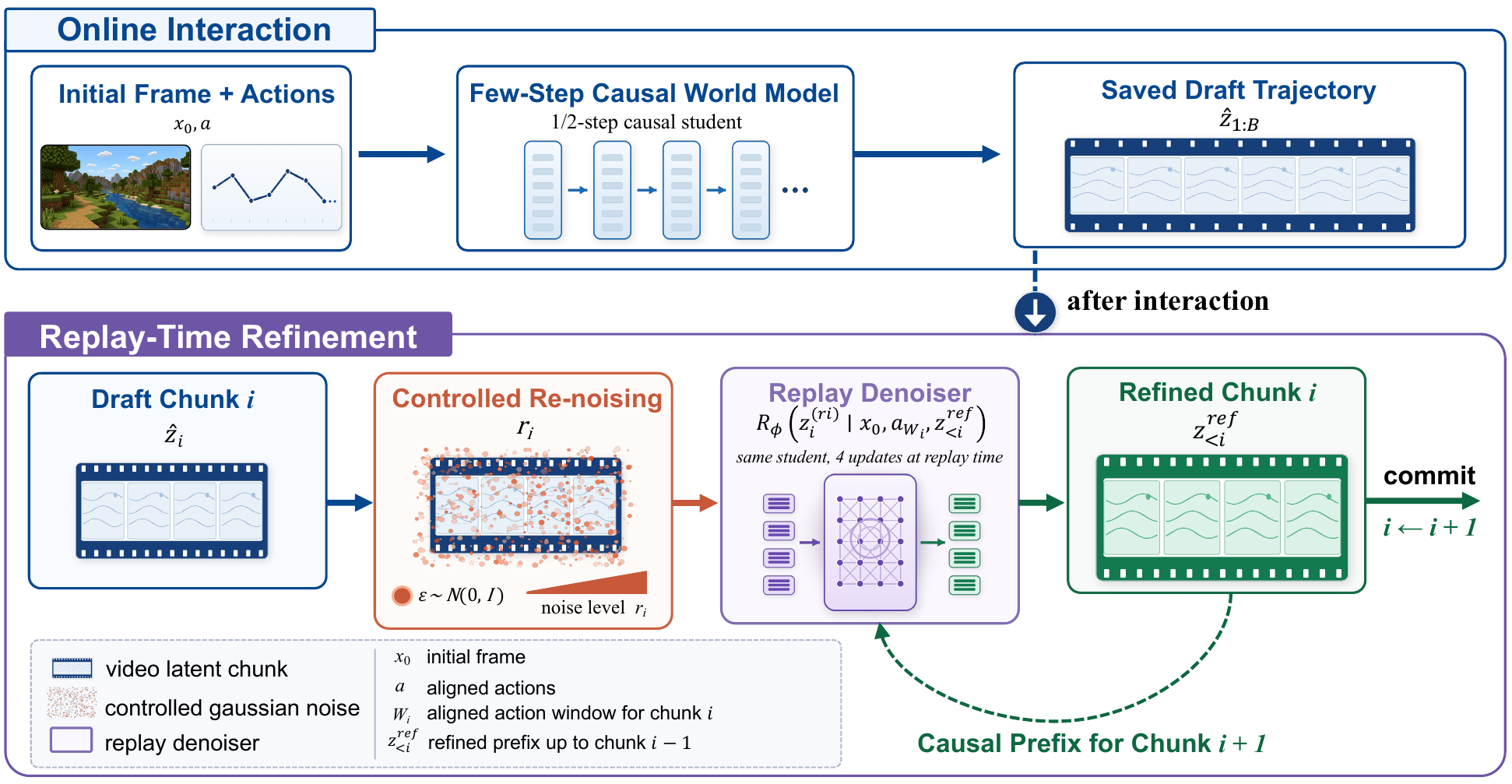}
    \caption{\textbf{Replay-Time Refinement.} The deployed student re-noises
    and denoises its own saved chunks under aligned actions and the refined
    causal prefix; each refined chunk is committed before the next is
    processed.}
    \label{fig:replay-refinement}
\end{figure}
 
Given a saved rollout $\hat z_{1:B}$, we retain its initial observation,
actions, and latent chunks. Figure~\ref{fig:replay-refinement} summarizes
the resulting sequential procedure. Each draft chunk is re-noised at an
intermediate flow time $r_i$:
\begin{equation}
    z_i^{(r_i)}=(1-r_i)\hat z_i+r_i\epsilon_i,
    \qquad \epsilon_i\sim\mathcal N(0,I).
\end{equation}
The deployed student then denoises it under the recorded action window and the
already-refined causal prefix:
\begin{equation}
    z_i^{\mathrm{ref}}
    =\mathcal R_{\phi}^{\mathcal S(r_i)}
    \left(z_i^{(r_i)};x_0,a_{\mathcal W_i},
    z_{<i}^{\mathrm{ref}}\right).
\end{equation}
In the default setting, $\mathcal R_\phi$ is the frozen ForgeWM-1 draft
model, and $\mathcal S(r_i)$ consists of four updates from $r_i=0.3$ to
zero. Replay therefore requires no parameter updates or separately trained
refiner.
Each refined chunk is committed before processing the next. The noise level
$r_i$ controls the trade-off between draft retention and visual revision.
Because replay occurs after interaction, it adds no online computation.

\begin{figure}[t]
    \centering

    \begin{minipage}[t]{0.57\textwidth}
        \vspace{0pt}
        \centering

        \includegraphics[width=\linewidth]
            {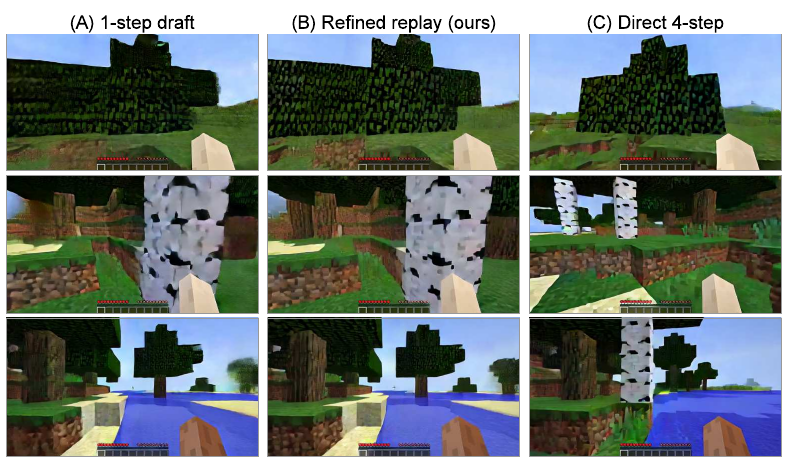}

        \caption{\textbf{Replay refinement versus direct generation.}
        Each row shares the same initial frame and controls.
        \textbf{(A)} Saved ForgeWM-1 draft;
        \textbf{(B)} replay-refined result;
        \textbf{(C)} direct ForgeWM-4 generation from noise.}
        \label{fig:replay-qualitative}
    \end{minipage}
    \hfill
    \begin{minipage}[t]{0.41\textwidth}
    \vspace{0pt}
        \vspace*{0.9\baselineskip}

    \centering
    \emergencystretch=1em

    {
    \small
    \setlength{\tabcolsep}{2.5pt}
    \renewcommand{\arraystretch}{1.18}

    \begin{tabularx}{\linewidth}{
        @{}
        >{\raggedright\arraybackslash}X
        c
        c
        @{}
    }
        \toprule
        Method
            & \shortstack{Replay\\LPIPS$\downarrow$}
            & $D_{\mathrm{draft}}\downarrow$ \\
        \midrule
        \multicolumn{3}{@{}l}{
            \textit{Draft-preserving methods}
        } \\
        One-step draft (no refinement)
            & 0.6532 & \textemdash \\
        \textbf{Replay refinement (ours)}
            & 0.6155 & 0.1970 \\
        \quad w/ four-step companion refiner
            & 0.6157 & 0.1960 \\
        \quad w/ four-step draft and refiner$^\dagger$
            & 0.6099 & 0.2361 \\
        \midrule
        \multicolumn{3}{@{}l}{
            \textit{From-noise baselines}
        } \\
        Direct ForgeWM-4 (from noise)
            & 0.6168 & 0.6187 \\
        \bottomrule
    \end{tabularx}
    }

    \captionof{table}{\textbf{Replay-Time Refinement.}
    Replay LPIPS is computed against the paired reference;
    $D_{\mathrm{draft}}$ is LPIPS to the saved draft.
    The $^\dagger$ row is evaluated against its own four-step
    draft and is therefore not directly comparable with the
    remaining rows.}
    \label{tab:replay-refinement}
\end{minipage}

\end{figure}
\paragraph{Replay results.}
Figure~\ref{fig:replay-qualitative} illustrates the distinction between
refining a realized rollout and resampling a new one. Starting from the
one-step draft in \textbf{(A)}, replay in \textbf{(B)} refines local visual
detail while preserving the draft's viewpoint and scene layout. Direct
ForgeWM-4 generation in \textbf{(C)} also produces a high-quality rollout,
but its fresh-noise initialization can lead to a different realization,
visibly changing the viewpoint and object layout.
Table~\ref{tab:replay-refinement} quantifies this distinction: replay achieves
an LPIPS of 0.6155, comparable to 0.6168 for direct ForgeWM-4, while reducing
$D_{\mathrm{draft}}$ from 0.6187 to 0.1970. Using the four-step companion as
the refiner yields nearly the same trade-off, indicating that replay does not
require a separate checkpoint.

\subsection{Recipe Transfer to FPS Gameplay}
\label{sec:crossfps}

To test whether the training recipe extends to a different control space, we apply the same
four-stage progression to first-person-shooter (FPS) gameplay. This produces
ForgeWM-CrossFPS, a separately trained checkpoint.  We adopt the cross-game FPS setting of
SCOPE~\cite{tong2026scope}, in
which controls are gamepad signals: two continuous analog sticks for movement
and look, together with discrete action buttons (fire, aim, jump, reload,
weapon switch, and melee).  Using a similarly structured action interface and 4-stage
recipe without architectural changes yields a four-step causal student.

\begin{figure}[!t]
    \centering
    \includegraphics[width=\textwidth]{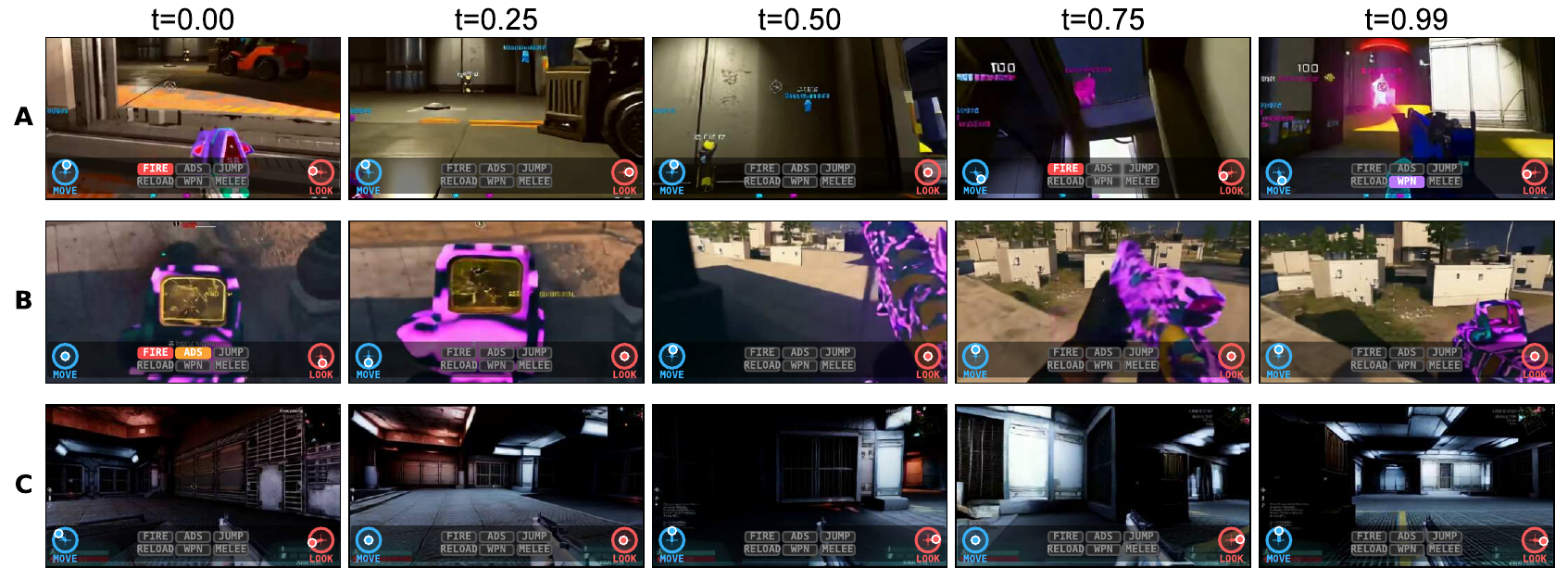}
    \caption{\textbf{CrossFPS rollouts.} ForgeWM-CrossFPS on
    \textbf{(A)}~Halo Infinite, \textbf{(B)}~Modern Warfare, and
    \textbf{(C)}~a science-fiction shooter.  Colored overlays indicate active
    gamepad controls.}
    \label{fig:crossfps-showcase}
\end{figure}

\FloatBarrier

Figure~\ref{fig:crossfps-showcase} shows representative rollouts from three games.  The macro-average
paired LPIPS is $0.656$, while the generated-to-reference motion ratio averages
$1.45$, suggesting a tendency toward stronger motion than the reference.  These measurements use a
different domain and protocol from the Minecraft comparison;
Appendix~\ref{app:crossfps} reports the per-game breakdown.

\subsection{Extended Qualitative Rollouts}
\label{sec:long-rollouts}

\begin{figure}[!h]
    \centering
    \includegraphics[width=\textwidth]{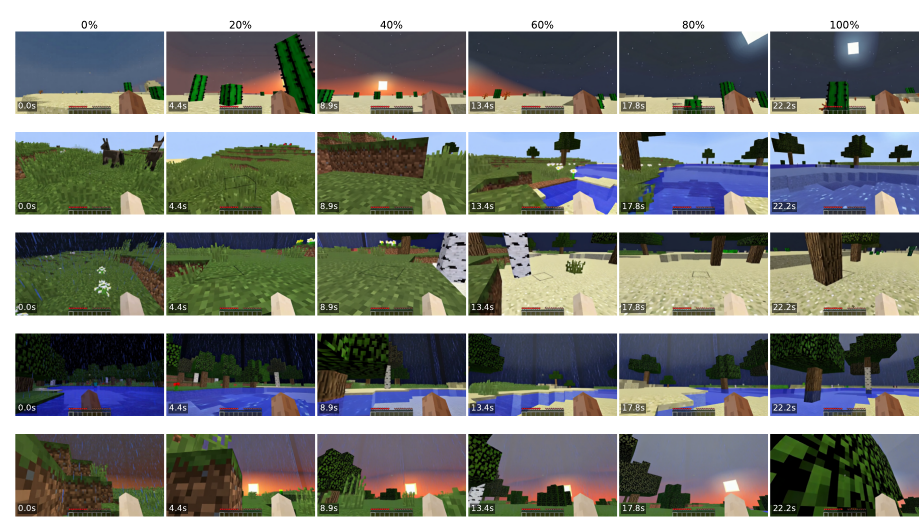}
    \caption{\textbf{Extended qualitative rollouts} from the four-step
    Minecraft checkpoint.  Each row shows one $22$\,s causal rollout sampled at
    evenly spaced timestamps, with elapsed time marked per frame.  Rows span
    biome and time of day and include both sustained forward motion and
    sustained turning.}
    \label{fig:long-rollouts}
\end{figure}

Figure~\ref{fig:long-rollouts} shows five $22$\,s rollouts, roughly
$3.5\times$ the horizon used for the quantitative protocol, selected to cover
different biomes, lighting conditions, and control patterns.  Some sequences
exhibit a gradual loss of block structure or slowly spreading color artifacts
at later timestamps, illustrating the remaining long-horizon degradation modes
(Appendix~\ref{app:limitations}).

\subsection{Evaluation Scope}

Our primary evaluation intentionally focuses on a controlled Minecraft
setting, where models can be assessed from matched initial states and mapped
control traces. Broader out-of-distribution generalization is therefore beyond
the scope of the current study. Because HY-WorldPlay uses a different control
parameterization, its effective motion scale depends on the deterministic
adapter. We consequently restrict direct quantitative comparisons to
Minecraft and present CrossFPS separately as a complementary evaluation of
recipe transfer to another gameplay domain.

\section{Conclusion}
\label{sec:conclusion}

We presented ForgeWM, a progressive causal training framework for
budget-specialized few-step world models that preserves frame-aligned
game-native controls.  Across its
budget-specialized students, ForgeWM delivers the strongest overall
quality--control profile among the evaluated systems on paired Minecraft
trajectories.
ForgeWM further exposes
distinct 1-, 2-, and 4-step operating points and a replay-refinement path that
separates latency-critical interaction from optional quality-oriented
processing, where the deployed one-step student refines its own saved draft and
matches four-step reference quality without a second checkpoint.  The same
training procedure also transfers to a gamepad-controlled FPS setting, while
our controlled quantitative comparison remains on Minecraft.  Together, these
components provide an effective training and deployment framework for
controllable few-step video world models.

\clearpage
\bibliographystyle{plainnat}
\bibliography{references}

@inproceedings{alonso2024diamond,
 author = {Alonso, Eloi and Jelley, Adam and Micheli, Vincent and Kanervisto, Anssi and Storkey, Amos and Pearce, Tim and Fleuret, Fran\c{c}ois},
 booktitle = {Advances in Neural Information Processing Systems},
 doi = {10.52202/079017-1873},
 editor = {A. Globerson and L. Mackey and D. Belgrave and A. Fan and U. Paquet and J. Tomczak and C. Zhang},
 pages = {58757--58791},
 publisher = {Curran Associates, Inc.},
 title = {Diffusion for World Modeling: Visual Details Matter in Atari},
 url = {https://proceedings.neurips.cc/paper_files/paper/2024/file/6bdde0373d53d4a501249547084bed43-Paper-Conference.pdf},
 volume = {37},
 year = {2024}
}

@inproceedings{valevski2024gamengen,
  title     = {Diffusion Models Are Real-Time Game Engines},
  author    = {Valevski, Dani and Leviathan, Yaniv and Arar, Moab and Fruchter, Shlomi},
  booktitle = {International Conference on Learning Representations},
  year      = {2025}
}

@article{tong2026scope,
  title   = {{SCOPE}: Simulating Cross-game Operations in Playable Environments for {FPS} World Models},
  author  = {Tong, Zizhao and Jin, Yeying and Lai, Hongfeng and Wang, Zeqing and Xing, Zhaohu and Cheng, Kexu and Xu, Haoran and Pu, Zhao and Zhu, Shangwen and Feng, Ruili and Zhao, Jian and Zhang, Yan and Tang, Hao and Shao, Ling},
  journal = {arXiv preprint arXiv:2605.23345},
  year    = {2026}
}

@inproceedings{chen2024diffusionforcing,
 author = {Chen, Boyuan and Mart\'{\i} Mons\'{o}, Diego and Du, Yilun and Simchowitz, Max and Tedrake, Russ and Sitzmann, Vincent},
 booktitle = {Advances in Neural Information Processing Systems},
 doi = {10.52202/079017-0759},
 editor = {A. Globerson and L. Mackey and D. Belgrave and A. Fan and U. Paquet and J. Tomczak and C. Zhang},
 pages = {24081--24125},
 publisher = {Curran Associates, Inc.},
 title = {Diffusion Forcing: Next-token Prediction Meets Full-Sequence Diffusion},
 url = {https://proceedings.neurips.cc/paper_files/paper/2024/file/2aee1c4159e48407d68fe16ae8e6e49e-Paper-Conference.pdf},
 volume = {37},
 year = {2024}
}

@article{lin2025aapt,
  title   = {Autoregressive Adversarial Post-Training for Real-Time Interactive Video Generation},
  author  = {Lin, Shanchuan and Yang, Ceyuan and He, Hao and Jiang, Jianwen and Ren, Yuxi and Xia, Xin and Zhao, Yang and Xiao, Xuefeng and Jiang, Lu},
  journal = {arXiv preprint arXiv:2506.09350},
  year    = {2025}
}

@article{xiang2026ttc,
  title   = {Pathwise Test-Time Correction for Autoregressive Long Video Generation},
  author  = {Xiang, Xunzhi and Duan, Zixuan and Zhang, Guiyu and Zhang, Haiyu and Gao, Zhe and Wu, Junta and Zhang, Shaofeng and Wang, Tengfei and Fan, Qi and Guo, Chunchao},
  journal = {arXiv preprint arXiv:2602.05871},
  year    = {2026}
}

@article{he2025matrixgame2,
  title={Matrix-Game 2.0: An Open-Source, Real-Time, and Streaming Interactive World Model},
  author={He, Xianglong and Peng, Chunli and Liu, Zexiang and Wang, Boyang and Zhang, Yifan and Cui, Qi and Kang, Fei and Jiang, Biao and An, Mengyin and Ren, Yangyang and Xu, Baixin and Guo, Hao-Xiang and Gong, Kaixiong and Wu, Cyrus and Li, Wei and Song, Xuchen and Liu, Yang and Li, Eric and Zhou, Yahui},
  journal={arXiv preprint arXiv:2508.13009},
  year={2025}
}

@article{wang2026matrixgame3,
  title   = {Matrix-Game 3.0: Real-Time and Streaming Interactive World Model with Long-Horizon Memory},
  author  = {Wang, Zile and Liu, Zexiang and Li, Jiaxing and Huang, Kaichen and Xu, Baixin and Kang, Fei and An, Mengyin and Wang, Peiyu and Jiang, Biao and Wei, Yichen and Xietian, Yidan and Pei, Jiangbo and Hu, Liang and Jiang, Boyi and Xue, Hua and Wang, Zidong and Sun, Haofeng and Li, Wei and Ouyang, Wanli and He, Xianglong and Liu, Yang and Li, Yangguang and Zhou, Yahui},
  journal = {arXiv preprint arXiv:2604.08995},
  year    = {2026}
}

@article{zhao2026minwm,
  title   = {minWM: A Full-Stack Open-Source Framework for Real-Time Interactive Video World Models},
  author  = {Zhao, Min and Zhu, Hongzhou and Yan, Bokai and Zhou, Zihan and Chen, Yimin and Sun, Wenqiang and Zheng, Kaiwen and He, Guande and Yang, Xiao and Li, Chongxuan and Bao, Fan and Zhu, Jun},
  journal = {arXiv preprint arXiv:2605.30263},
  year    = {2026}
}

@inproceedings{li2025prope,
 author = {Li, Ruilong and Yi, Brent and Liu, Junchen and Gao, Hang and Ma, Yi and Kanazawa, Angjoo},
 booktitle = {Advances in Neural Information Processing Systems},
 editor = {D. Belgrave and C. Zhang and H. Lin and R. Pascanu and P. Koniusz and M. Ghassemi and N. Chen},
 pages = {15984--16009},
 publisher = {Curran Associates, Inc.},
 title = {Cameras as Relative Positional Encoding},
 url = {https://proceedings.neurips.cc/paper_files/paper/2025/file/17a7075094632c88cccdd86270ad715b-Paper-Conference.pdf},
 volume = {38},
 year = {2025}
}

@inproceedings{huang2025selfforcing,
  title     = {Self Forcing: Bridging the Train-Test Gap in Autoregressive Video Diffusion},
  author    = {Huang, Xun and Li, Zhengqi and He, Guande and Zhou, Mingyuan and Shechtman, Eli},
  booktitle = {Advances in Neural Information Processing Systems},
  year      = {2025}
}

@inproceedings{zhu2026causalforcing,
  title     = {Causal Forcing: Autoregressive Diffusion Distillation Done Right for High-Quality Real-Time Interactive Video Generation},
  author    = {Zhu, Hongzhou and Zhao, Min and He, Guande and Su, Hang and Li, Chongxuan and Zhu, Jun},
  booktitle = {International Conference on Machine Learning},
  year      = {2026}
}

@article{zhao2026causalforcingpp,
  title   = {Causal Forcing++: Scalable Few-Step Autoregressive Diffusion Distillation for Real-Time Interactive Video Generation},
  author  = {Zhao, Min and Zhu, Hongzhou and Zheng, Kaiwen and Zhou, Zihan and Yan, Bokai and Li, Xinyuan and Yang, Xiao and Li, Chongxuan and Zhu, Jun},
  journal = {arXiv preprint arXiv:2605.15141},
  year    = {2026}
}

@article{zheng2026causalrcm,
  title   = {Causal-rCM: A Unified Teacher-Forcing and Self-Forcing Open Recipe for Autoregressive Diffusion Distillation in Streaming Video Generation and Interactive World Models},
  author  = {Zheng, Kaiwen and He, Guande and Zhao, Min and Zhang, Jintao and Chen, Huayu and Chen, Jianfei and Lin, Chen-Hsuan and Liu, Ming-Yu and Zhu, Jun and Ma, Qianli},
  journal = {arXiv preprint arXiv:2606.25473},
  year    = {2026}
}

@article{gu2026anyflow,
  title   = {AnyFlow: Any-Step Video Diffusion Model with On-Policy Flow Map Distillation},
  author  = {Gu, Yuchao and Fang, Guian and Jiang, Yuxin and Mao, Weijia and Han, Song and Cai, Han and Shou, Mike Zheng},
  journal = {arXiv preprint arXiv:2605.13724},
  year    = {2026}
}

@inproceedings{meng2021sdedit,
  title     = {SDEdit: Guided Image Synthesis and Editing with Stochastic Differential Equations},
  author    = {Meng, Chenlin and He, Yutong and Song, Yang and Song, Jiaming and Wu, Jiajun and Zhu, Jun-Yan and Ermon, Stefano},
  booktitle = {International Conference on Learning Representations},
  year      = {2022}
}

@article{yu2025autorefiner,
  title   = {AutoRefiner: Improving Autoregressive Video Diffusion Models via Reflective Refinement Over the Stochastic Sampling Path},
  author  = {Yu, Zhengyang and Hayakawa, Akio and Ishii, Masato and Yu, Qingtao and Shibuya, Takashi and Zhang, Jing and Mitsufuji, Yuki},
  journal = {arXiv preprint arXiv:2512.11203},
  year    = {2025}
}

@article{zhu2026sanawm,
  title   = {SANA-WM: Efficient Minute-Scale World Modeling with Hybrid Linear Diffusion Transformer},
  author  = {Zhu, Haoyi and Liu, Haozhe and Zhao, Yuyang and Ye, Tian and Chen, Junsong and Yu, Jincheng and He, Tong and Han, Song and Xie, Enze},
  journal = {arXiv preprint arXiv:2605.15178},
  year    = {2026}
}

@article{worldplay2025,
  title   = {WorldPlay: Towards Long-Term Geometric Consistency for Real-Time Interactive World Modeling},
  author  = {Sun, Wenqiang and Zhang, Haiyu and Wang, Haoyuan and Wu, Junta and Wang, Zehan and Wang, Zhenwei and Wang, Yunhong and Zhang, Jun and Wang, Tengfei and Guo, Chunchao},
  journal = {arXiv preprint arXiv:2512.14614},
  year    = {2025}
}

@inproceedings{huang2023vbench,
  title     = {{VBench}: Comprehensive Benchmark Suite for Video Generative Models},
  author    = {Huang, Ziqi and He, Yinan and Yu, Jiashuo and Zhang, Fan and Si, Chenyang and Jiang, Yuming and Zhang, Yuanhan and Wu, Tianxing and Jin, Qingyang and Chanpaisit, Nattapol and Wang, Yaohui and Chen, Xinyuan and Wang, Limin and Lin, Dahua and Qiao, Yu and Liu, Ziwei},
  booktitle = {Proceedings of the IEEE/CVF Conference on Computer Vision and Pattern Recognition},
  year      = {2024}
}

@inproceedings{yu2025gamefactory,
  title     = {GameFactory: Creating New Games with Generative Interactive Videos},
  author    = {Yu, Jiwen and Qin, Yiran and Wang, Xintao and Wan, Pengfei and Zhang, Di and Liu, Xihui},
  booktitle = {Proceedings of the IEEE/CVF International Conference on Computer Vision},
  pages     = {11590--11599},
  year      = {2025}
}

@inproceedings{yang2025asd,
title={Towards One-step Causal Video Generation via Adversarial Self-Distillation},
author={Yongqi Yang and Huayang Huang and Xu Peng and Xiaobin Hu and Donghao Luo and Jiangning Zhang and Chengjie Wang and Yu Wu},
booktitle={The Fourteenth International Conference on Learning Representations},
year={2026},
url={https://openreview.net/forum?id=P3O0fNmnWa}
}

@inproceedings{lipman2022flow,
  title={Flow matching for generative modeling},
  author={Lipman, Yaron and Chen, Ricky TQ and Ben-Hamu, Heli and Nickel, Maximilian and Le, Matthew},
  booktitle={The eleventh international conference on learning representations},
  year= {2023}
}

@inproceedings{song2023consistency,
  title     = {Consistency Models},
  author    = {Song, Yang and Dhariwal, Prafulla and Chen, Mark and Sutskever, Ilya},
  booktitle = {Proceedings of the 40th International Conference on Machine Learning},
  year      = {2023}
}

@inproceedings{yin2024dmd,
 author = {Yin, Tianwei and Gharbi, Micha{\"e}l and Zhang, Richard and Shechtman, Eli and Durand, Fredo and Freeman, William T. and Park, Taesung},
 booktitle = {Proceedings of the IEEE/CVF Conference on Computer Vision and Pattern Recognition},
 title = {One-step Diffusion with Distribution Matching Distillation},
 year = {2024}
}

@inproceedings{zhang2018lpips,
 author = {Richard Zhang and Phillip Isola and Alexei A. Efros and Eli Shechtman and Oliver Wang},
 booktitle = {Proceedings of the IEEE Conference on Computer Vision and Pattern Recognition},
 journal = {2018 IEEE/CVF Conference on Computer Vision and Pattern Recognition},
 title = {The Unreasonable Effectiveness of Deep Features as a Perceptual Metric},
 year = {2018}
}

@inproceedings{bruce2024genie,
 author = {Bruce, Jake and Dennis, Michael and Edwards, Ashley and Parker-Holder, Jack and Shi, Yuge and Hughes, Edward and Lai, Matthew and Mavalankar, Aditi and Steigerwald, Richie and Apps, Chris and Aytar, Yusuf and Bechtle, Sarah and Behbahani, Feryal and Chan, Stephanie and Heess, Nicolas and Gonzalez, Lucy and Osindero, Simon and Ozair, Sherjil and Reed, Scott and Zhang, Jingwei and Zolna, Konrad and Clune, Jeff and de Freitas, Nando and Singh, Satinder and Rockt{\"a}schel, Tim},
 booktitle = {Proceedings of the 41st International Conference on Machine Learning},
 pages = {4603--4623},
 title = {Genie: Generative Interactive Environments},
 year = {2024}
}

@article{guo2025mineworld,
  title={MineWorld: a Real-Time and Open-Source Interactive World Model on Minecraft}, 
  author={Guo, Junliang and Ye, Yang and He, Tianyu and Wu, Haoyu and Jiang, Yushu and Pearce, Tim and Bian, Jiang},
  year={2025},
  journal={arXiv preprint arXiv:2504.08388},
}

@article{zhang2025matrixgame,
  title     = {Matrix-Game: Interactive World Foundation Model},
  author    = {Yifan Zhang and Chunli Peng and Boyang Wang and Puyi Wang and Qingcheng Zhu and Fei Kang and Biao Jiang and Zedong Gao and Eric Li and Yang Liu and Yahui Zhou},
  journal   = {arXiv preprint arXiv:2506.18701},
  year      = {2025}
}

@misc{nvidia2026cosmos3,
      title={Cosmos 3: Omnimodal World Models for Physical AI}, 
      author={{NVIDIA} and Aditi and Niket Agarwal and Arslan Ali and Jon Allen and Martin Antolini and Adeline Aubame and Alisson Azzolini and Junjie Bai and Maciej Bala and Yogesh Balaji and Josh Bapst and Aarti Basant and Mukesh Beladiya and Mohammad Qazim Bhat and Zaid Pervaiz Bhat and Dan Blick and Vanni Brighella and Han Cai and Tiffany Cai and Eric Cameracci and Jiaxin Cao and Yulong Cao and Mark Carlson and Carlos Casanova and Ting-Yun Chang and Yan Chang and Yu-Wei Chao and Prithvijit Chattopadhyay and Roshan Chaudhari and Chieh-Yun Chen and Junyu Chen and Ke Chen and Qizhi Chen and Wenkai Chen and Xiaotong Chen and Yu Chen and An-Chieh Cheng and Click Cheng and Xiu Chia and Jeana Choi and Chaeyeon Chung and Wenyan Cong and Yin Cui and Magdalena Dadela and Nalin Dadhich and Wenliang Dai and Joyjit Daw and Alperen Degirmenci and Rodrigo Vieira Del Monte and Robert Denomme and Sameer Dharur and Marco Di Lucca and Ke Ding and Wenhao Ding and Yifan Ding and Yuzhu Dong and Nicole Drumheller and Yilun Du and Aigul Dzhumamuratova and Aleksandr Efitorov and Hamid Eghbalzadeh and Naomi Eigbe and Imad El Hanafi and Hassan Eslami and Benedikt Falk and Jiaojiao Fan and Jim Fan and Amol Fasale and Sergiy Fefilatyev and Liang Feng and Francesco Ferroni and Sanja Fidler and Xiao Fu and Vikram Fugro and Prashant Gaikwad and TJ Galda and Katelyn Gao and Yihuai Gao and Wenhang Ge and Sreyan Ghosh and Arushi Goel and Vivek Goel and Akash Gokul and Rama Govindaraju and Jinwei Gu and Miguel Guerrero and Elfie Guo and Aryaman Gupta and Siddharth Gururani and Hugo Hadfield and Song Han and Ankur Handa and Zekun Hao and Mohammad Harrim and Ali Hassani and Nathan Hayes-Roth and Yufan He and Chris Helvig and Cyrus Hogg and Madison Huang and Michael Huang and Sophia Huang and Yufan Huang and Jacob Huffman and DeLesley Hutchins and Suneel Indupuru and Boris Ivanovic and Arihant Jain and Joel Jang and Ryan Ji and Yanan Jian and Dongfu Jiang and Jingyi Jin and Atharva Joshi and Nikhilesh Joshi and Pranjali Joshi and Andy Ju and Jaehun Jung and Weiwei Kang and Scott Kassekert and Jan Kautz and Ashna Khetan and Julia Kiczka and Slawek Kierat and Gwanghyun Kim and Kuno Kim and Sunny Kim and Kezhi Kong and Xin Kong and Zhifeng Kong and Tomasz Kornuta and Egor Krivov and Hui Kuang and Saurav Kumar and Chia-Wen Kuo and George Kurian and Wojciech Kutak and JF Lafleche and Himangshu Lahkar and Omar Laymoun and Jayjun Lee and Sanggil Lee and Gabriele Leone and Boyi Li and Freya Li and Jiajun Li and Jinfeng Li and Ling Li and Pengcheng Li and Shangru Li and Tingle Li and Xiaolong Li and Xuan Li and Zhaoshuo Li and Zhiqi Li and Hao Liang and Maosheng Liao and Chen-Hsuan Lin and Tsung-Yi Lin and Ming-Yu Liu and Sifei Liu and Zihan Liu and Hai Loc Lu and Xiangyu Lu and Alice Luo and Ruipu Luo and Wenjie Luo and Jiangran Lyu and Martin Ding Ma and Nic Ma and Qianli Ma and Dawid Majchrowski and Louis Marcoux and Miguel Martin and Qing Miao and Ashkan Mirzaei and Shreyas Misra and Kaichun Mo and Durra Mohsin and Hyejin Moon and Pawel Morkisz and Saeid Motiian and Kirill Motkov and Seungjun Nah and Yashraj Narang and Deepak Narayanan and Thabang Ngazimbi and Julian Ouyang and Shubham Pachori and David Page and Yatian Pang and Sehwi Park and Mahesh Patekar and Mostofa Patwary and Marco Pavone and Trung Pham and Wei Ping and Soha Pouya and Shrimai Prabhumoye and Varun Praveen and Delin Qu and Hesam Rabeti and Morteza Ramezanali and Marilyn Reeb and Xuanchi Ren and Kristen Rumley and Wojciech Rymer and Jun Saito and Yeongho Seol and John Shao and Piyush Shekdar and Tianwei Shen and Humphrey Shi and Min Shi and Stella Shi and Kevin Shih and Mohammad Shoeybi and Mateusz Sieniawski and Shuran Song and Alexander Sotelo and Amir Sotoodeh and Sunil Srinivasa and Vignesh Srinivasakumar and Bartosz Stefaniak and Rahul Heinrich Steiger and Shangkun Sun and Jiaxiang Tang and Shitao Tang and Yangyang Tang and Yue Tang and Tolou Tavakkoli and Kayley Ting and Krzysztof Tomala and Wei-Cheng Tseng and Jibin Varghese and Sergei Vasilev and Thomas Volk and Raju Wagwani and Roger Waleffe and Andrew Z. Wang and Boxiang Wang and Haoxiang Wang and Qiao Wang and Shihao Wang and Shijie Wang and Ting-Chun Wang and Yan Wang and Yu Wang and Rohit Watve and David Wehr and Fangyin Wei and Xinshuo Weng and Jay Zhangjie Wu and Kedi Wu and Hongchi Xia and Summer Xiao and Tianjun Xiao and Kevin Xie and Daguang Xu and Jiashu Xu and Mengyao Xu and Ruqing Xu and Xingqian Xu and Yao Xu and Dinghao Yang and Dong Yang and Hans Yang and Xiaodong Yang and Xuning Yang and Yichu Yang and Yurong You and Zhiding Yu and Hao Yuan and Simon Yuen and Xiaohui Zeng and Pengcuo Zeren and Cindy Zha and Haotian Zhang and Jenny Zhang and Jing Zhang and Liangkai Zhang and Paris Zhang and Shun Zhang and Xuanmeng Zhang and Zhizheng Zhang and Ann Zhao and Yilin Zhao and Yuliya Zhautouskaya and Charles Zhou and Fengzhe Zhou and Shilin Zhu and Yuke Zhu and Dima Zhylko and Artur Zolkowski},
      year={2026},
      eprint={2606.02800},
      archivePrefix={arXiv},
      primaryClass={cs.CV},
      url={https://arxiv.org/abs/2606.02800}, 
}

@article{gao2026lingbotworld2,
  title={Infinite Worlds with Versatile Interactions}, 
  author={Zelin Gao and Qiuyu Wang and Jiapeng Zhu and Jingye Chen and Zichen Liu and Qingyan Bai and Jiahao Wang and Yufeng Yuan and Hanlin Wang and Yichong Lu and Ka Leong Cheng and Haojie Zhang and Jian Gao and Tianrui Feng and Yuzheng Liu and Yao Yao and Yinghao Xu and Xing Zhu and Yujun Shen and Hao Ouyang},
  journal={arXiv preprint arXiv:2607.07534},
  year={2026}
}

@article{dreamx2026world,
  title={DreamX-World 1.0: A General-Purpose Interactive World Model},
  author={{DreamX Team} and Bai, Yancheng and Chen, Rui and Chu, Xiangxiang and Dang, Rujing and Dou, Hao and Gao, Bingjie and Gu, Qiwen and Hong, Siyu and Lei, Jiachen and others},
  journal={arXiv preprint arXiv:2606.16993},
  year={2026}
}

\beginappendix

\makeatletter
\@addtoreset{figure}{section}
\@addtoreset{table}{section}
\makeatother

\renewcommand{\thefigure}{\Alph{section}\arabic{figure}}
\renewcommand{\thetable}{\Alph{section}\arabic{table}}
\section{Training Details}
\label{app:training-details}

\FloatBarrier
\subsection{Optimization and Stage-Wise Settings}

\paragraph{Shared setup.}
All stages train the Matrix-Game 2.0 image-to-video
lineage~\cite{he2025matrixgame2} (a Wan2.1-T2V-1.3B backbone with
our action-conditioning module) on GF-Minecraft
clips~\cite{yu2025gamefactory} at $640\!\times\!352$ and $12$ frames per
second. A VAE with $4\times$ temporal compression maps each clip to a latent
sequence of $21$ latent frames ($16\!\times\!44\!\times\!80$), so a
three-latent causal chunk spans twelve video frames. We optimize with AdamW
($\beta=(0.0,0.999)$), mixed-precision \texttt{bf16}, gradient checkpointing,
and fully-sharded data parallelism across eight GPUs; the seed is fixed at $0$.
The flow-matching path uses $1000$ discretization steps with a timestep shift
of $5.0$; the consistency and distribution-matching stages apply
classifier-free guidance with weight $3.0$ to the frozen teacher. Learning
rates are held constant within each stage.

\begin{table}[h]
    \centering
    {
    \small
    \setlength{\tabcolsep}{3pt}
    \begin{tabular*}{\textwidth}{@{\extracolsep{\fill}}lcccc@{}}
        \toprule
        & \textbf{Stage 0} & \textbf{Stage 1} & \textbf{Stage 2}
        & \textbf{Stage 3} \\
        \midrule
        Objective        & $\mathcal L_{\mathrm{FM}}$ & $\mathcal L_{1}$
                         & $\mathcal L_{2}$ & $\mathcal L_{3}$ \\
        Trainer          & FM & causal FM & consist.\ distill.\ & DMD \\
        Attention        & bidir. & causal & causal & causal \\
        Init             & base & base & Stage 1 & Stage 2 \\
        Blk.\ (latents)  & 21 & 3 & 3 & 3 \\
        Trainer iterations          & 4k & 20k & 6k & 4k \\
        Gen.\ lr         & $2\mathrm{e}{-6}$ & $2\mathrm{e}{-5}$
                         & $2\mathrm{e}{-6}$ & $2\mathrm{e}{-6}$ \\
        Critic lr        & -- & -- & -- & $4\mathrm{e}{-7}$ \\
        Global batch     & 8 & 8 & 8 & 8 \\
        EMA              & -- & -- & $0.99$/200 & $0.99$/200 \\
        \bottomrule
    \end{tabular*}
    }
    \caption{\textbf{Per-stage optimization for the reported Minecraft ForgeWM
    lineage.} ``Blk.'' is the number of latent frames per attention block
    ($21$ = full-clip bidirectional; $3$ = one causal chunk). ``EMA'' is the
    generator exponential-moving-average decay and the step it starts (``--''
    when disabled). Global batch is the effective batch after data-parallel
    accumulation. All learning rates are constant.}
    \label{tab:training-hparams}
\end{table}

\begin{figure}[t]
    \centering
    \includegraphics[width=0.78\textwidth]{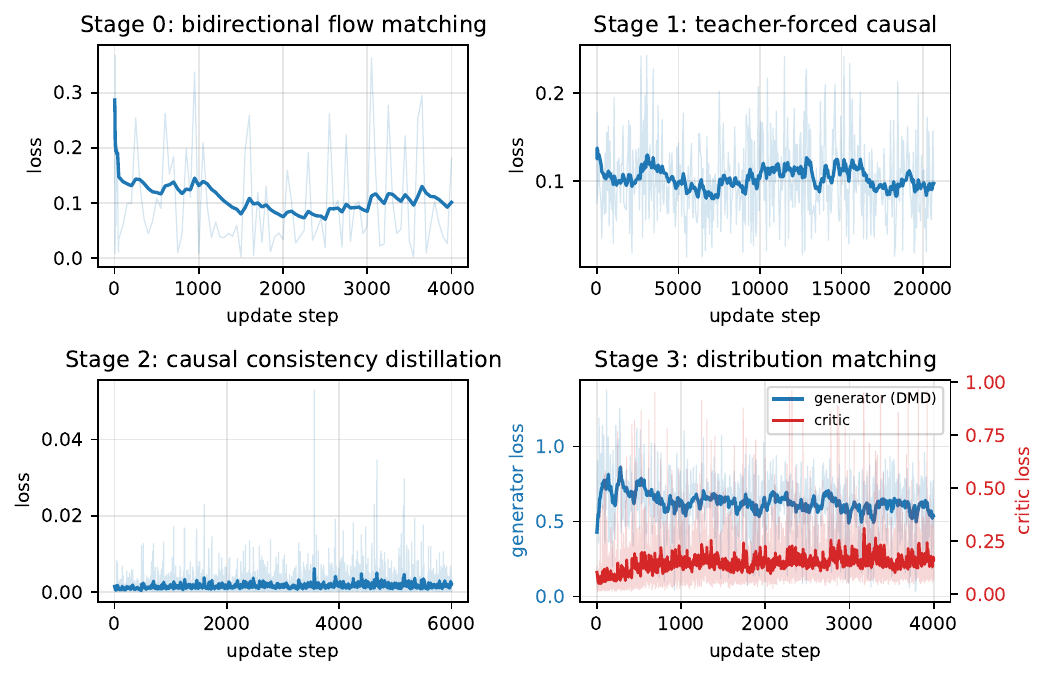}
    \caption{\textbf{Training losses across stages.}
    Raw values are shown as light traces, with exponential-moving-average
    trends overlaid; the horizontal axis denotes trainer iterations. In
    Stage~3, the critic is updated every iteration and the generator every
    fifth iteration.
    \emph{Top-left}: Stage~0 bidirectional flow matching.
    \emph{Top-right}: Stage~1 teacher-forced causal flow matching.
    \emph{Bottom-left}: Stage~2 online causal consistency distillation.
    \emph{Bottom-right}: Stage~3 distribution matching, with the generator
    (DMD) loss on the left axis and the critic (fake-denoiser) loss on the
    right axis. Because the stages optimize different objectives, absolute
    loss magnitudes are not comparable across panels; the curves document
    optimization behavior within each stage only.}
    \label{fig:training-loss}
\end{figure}

\paragraph{Stage-specific settings.}
Table~\ref{tab:training-hparams} lists the per-stage optimization.
\emph{Stage~0} adapts the base generator with full-clip bidirectional
attention over $21$-latent clips and no EMA; we freeze its $4{,}000$-update
checkpoint and reuse it as the real denoiser
$\hat z_{\mathrm{real}}$ in Stage~3. \emph{Stage~1} replaces full attention
with block-wise causal attention (three latent frames per block) and trains the
teacher-forced objective from the same base initialization; the reported
lineage uses the $20{,}000$-update checkpoint. \emph{Stage~2} initializes the
generator, its EMA copy, and the frozen teacher from that Stage~1 checkpoint and
runs online causal consistency distillation for $6{,}000$ updates over a
discrete grid of $N\!=\!48$ noise levels, sampling one adjacent pair per step.
\emph{Stage~3} initializes from the Stage~2 checkpoint and comprises
three separate $4{,}000$-trainer-iteration runs for
$K\!\in\!\{1,2,4\}$. The four-step student uses the denoising schedule
$[1000,750,500,250]$ for every generated chunk. Following
\citet{yang2025asd}, the one- and two-step students use the same fixed
four-step schedule for the first generated chunk and their respective
budget-matched $K$-step schedules thereafter. Each run uses a local
attention window of six latent frames, equivalent to two three-latent
chunks. The critic is updated at every trainer iteration, whereas the
generator is updated once every five iterations. Figure~\ref{fig:training-loss}
plots the per-stage training losses recorded during these runs. The
reported ForgeWM-1/2/4 models are the corresponding Stage~3
checkpoints.

The CrossFPS checkpoint reported in the main paper follows the same
sequence of four training objectives on gamepad-controlled data, using
the widened continuous-control interface described in
Appendix~\ref{app:crossfps-adaptation}.

\FloatBarrier
\subsection{Stage-Wise Inference Ablation}
\label{app:stage-ablation}

The body compares checkpoints from Stages~1--3 under a shared four-step
inference schedule; this section gives the protocol and results, and adds the
Stage~0 domain teacher as a non-causal reference point.

\paragraph{Protocol.}
For Stages~1--3, we hold the inference path fixed and vary only the loaded
generator weights. All three checkpoints use the same chunked causal pipeline,
model configuration, four-step schedule $[1000,750,500,250]$, unrestricted
causal attention used for the main-table results, and identical
per-trajectory noise seeds. Stage~0 is bidirectional and has no chunked
few-step deployment; we therefore evaluate it in its native regime, using a
single full-clip block and a four-step UniPC schedule. Its row serves as a
non-causal domain-teacher reference and is not directly comparable to the
causal rows. All rows are scored on the same $1{,}000$ trajectories using the
paired LPIPS protocol of Appendix~\ref{app:eval-protocol}; intervals are
trajectory-level bootstrap 95\% confidence intervals.

\begin{table}[t]
    \centering
    {
    \small
    \setlength{\tabcolsep}{3pt}
    \begin{tabular*}{\textwidth}{@{\extracolsep{\fill}}clcccc@{}}
        \toprule
        \textbf{Stage} & \textbf{Inference regime}
        & \textbf{LPIPS}$\downarrow$ & \textbf{IQ}$\uparrow$
        & \textbf{AQ}$\uparrow$ & \textbf{SC}$\uparrow$ \\
        \midrule
        0 & bidirectional teacher (ref.) & $0.814_{[.809,.819]}$ & $0.455$ & $0.463$ & $0.677$ \\
        1 & teacher-forced causal        & $0.806_{[.799,.812]}$ & $0.508$ & $0.454$ & $0.700$ \\
        2 & causal consistency           & $\mathbf{0.605}_{[.600,.610]}$ & $0.659$ & $0.483$ & $\mathbf{0.760}$ \\
        3 & distribution matching        & $0.617_{[.613,.620]}$ & $\mathbf{0.716}$ & $\mathbf{0.489}$ & $\mathbf{0.760}$ \\
        \bottomrule
    \end{tabular*}
    }
    \caption{\textbf{Stage-wise inference ablation on $1{,}000$ paired
trajectories.} LPIPS uses the AlexNet backbone on $16$ evaluated frames;
subscripts denote trajectory-level bootstrap 95\% confidence intervals.
IQ, AQ, and SC are the corresponding VBench metrics. Stage~0 is a non-causal
domain-teacher reference evaluated in its native full-clip regime.}
    \label{tab:stage-ablation}
\end{table}

\paragraph{Findings.}
Table~\ref{tab:stage-ablation} isolates the contribution of each training
stage under a shared four-step evaluation protocol. Stages~0 and~1 are trained
as full-trajectory flow-matching models and are not intended for four-step
sampling; under this budget, they obtain LPIPS scores of $0.814$ and $0.806$,
respectively. Stage~2 produces the largest improvement, reducing LPIPS from
$0.806$ to $0.605$ with non-overlapping confidence intervals. This result
shows that causal consistency distillation, rather than teacher-forced
causalization alone, is the stage that enables effective few-step sampling.

Stage~3 introduces training on self-generated autoregressive histories. At
four steps, it obtains an LPIPS of $0.617$, with Stage~2 retaining a small
paired advantage of $-0.012$ ($[-0.015,-0.008]$). Its effect is more visible
in the no-reference metrics: Imaging Quality increases from $0.659$ to
$0.716$, while Subject Consistency remains unchanged at $0.760$ and Aesthetic
Quality changes only slightly. Thus, at the four-step budget, Stage~3 shifts
the trade-off toward sharper per-frame appearance rather than improving paired
reconstruction fidelity. Its incremental effect at one- and two-step budgets
is not isolated by this ablation.

\section{Evaluation Protocol}
\label{app:eval-protocol}

\subsection{Action Conversion Across Control Interfaces}
\label{app:action-conversion}

The three compared systems do not expose the same control interface, so a single
recorded control trace is mapped onto each system's own action parameterization
and every model is then driven from that mapped trace.  The source trace is the
frame-rate stream used throughout this work: a discrete keyboard state and a
continuous two-dimensional mouse delta per video frame, recorded at
$640\!\times\!352$ and $12$ frames per second over the 77-frame comparison
window.  ForgeWM and Matrix-Game 2.0 share this parameterization, so for those
two the mapping is the identity up to the sign convention of the vertical mouse
axis, which differs between the source recording and the model interface.
HY-WorldPlay additionally accepts a continuous camera parameterization, and the
conversion for that pathway integrates the per-frame mouse deltas into the
angular quantities its interface expects.

\subsection{Metric Definitions}
\label{app:metric-definitions}

Table~\ref{tab:world-model-comparison} reports nine columns.  Let
$X_n=(x_{n,0},\ldots,x_{n,T-1})$ be generated clip $n$, let
$R_n=(r_{n,0},\ldots,r_{n,T-1})$ be its temporally aligned reference when a
reference is required, and let $N$ be the number of evaluated clips.  All
reported dataset scores first average within a clip and then across clips. IQ
and AQ use the $462$ constant-action clips ($77$ scenes times six actions),
whereas SC, LPIPS, and Flow Profile use the $1{,}000$ shared-action rollouts.
Every metric reads exactly the first $77$ frames.

\paragraph{VBench metrics.}
Imaging Quality (IQ), Aesthetic Quality (AQ), and Subject Consistency (SC) use
the standard VBench implementation~\cite{huang2023vbench}.  Denote the MUSIQ
predictor by $q_{\rm MUSIQ}$, the LAION aesthetic linear predictor by
$q_{\rm aes}$, the normalized CLIP ViT-L/14 image feature by $c(\cdot)$, and the
normalized DINO ViT-B/16 feature by $d(\cdot)$.  Their dataset-level scores are
\begin{equation}
 {\rm IQ}=\frac{1}{NT}\sum_{n=1}^{N}\sum_{t=0}^{T-1}
 \frac{q_{\rm MUSIQ}(x_{n,t})}{100},
\end{equation}
\begin{equation}
 {\rm AQ}=\frac{1}{NT}\sum_{n=1}^{N}\sum_{t=0}^{T-1}
 \frac{q_{\rm aes}(c(x_{n,t}))}{10}.
\end{equation}
For Subject Consistency, let $[z]_+=\max(0,z)$ and define the local and
first-frame similarities
\begin{equation}
 u_{n,t}=[d(x_{n,t-1})^\top d(x_{n,t})]_+,
 \qquad
 v_{n,t}=[d(x_{n,0})^\top d(x_{n,t})]_+.
\end{equation}
Then
\begin{equation}
 {\rm SC}=\frac{1}{2N(T-1)}\sum_{n=1}^{N}\sum_{t=1}^{T-1}
 (u_{n,t}+v_{n,t}).
\end{equation}
Thus, SC rewards both local frame-to-frame consistency and retention of the
first-frame subject appearance; it is not an identity-classification accuracy.

\paragraph{Reference-aligned visual and motion metrics.}
Let
\begin{equation}
 \mathcal I=\{1,6,11,\ldots,76\}
\end{equation}
be the $16$ evaluated frame indices, excluding the given first frame, and let
$\ell_{\rm Alex}$ denote LPIPS with the AlexNet
backbone~\cite{zhang2018lpips}. We report
\begin{equation}
 {\rm LPIPS}=\frac{1}{N|\mathcal I|}\sum_{n=1}^{N}
 \sum_{t\in\mathcal I}\ell_{\rm Alex}(x_{n,t},r_{n,t}),
\end{equation}
where lower is better.

For Flow Profile similarity, let
$F(y_t,y_{t+1})\in{\rm R}^{H\times W\times2}$ denote the fixed dense
optical-flow estimator. For a video $Y$, define its temporal motion-magnitude
profile by
\begin{equation}
 p_k(Y)=\frac{1}{|\Omega|}\sum_{(u,v)\in\Omega}
 \left\|F(\tilde Y_{4k},\tilde Y_{4(k+1)})_{u,v}\right\|_2,
\end{equation}
for $k=0,\ldots,18$, and let
$\mathbf p(Y)=(p_0(Y),\ldots,p_{18}(Y))$.
Here $\tilde Y_t$ is frame $t$ converted to grayscale and resized to
$160\!\times\!88$ using area interpolation. The spatial evaluation region is
$\Omega=\Omega_h\times\Omega_w$, where
$\Omega_h=\{\lfloor0.08H\rfloor,\ldots,\lfloor0.82H\rfloor-1\}$
and
$\Omega_w=\{\lfloor0.08W\rfloor,\ldots,\lfloor0.92W\rfloor-1\}$.
This region suppresses the image borders and the HUD-heavy lower portion.
We compute $F$ using OpenCV Farneback flow with pyramid scale $0.5$, four
pyramid levels, window size $21$, four iterations, polynomial neighborhood
size $7$, polynomial standard deviation $1.5$, and flags set to zero.

The reported score is the mean paired cosine similarity
\begin{equation}
 {\rm FlowProf}=\frac{1}{N}\sum_{n=1}^{N}
 \frac{\mathbf p(X_n)^\top\mathbf p(R_n)}
 {\|\mathbf p(X_n)\|_2\|\mathbf p(R_n)\|_2}.
\end{equation}
For each summand, the cosine score is defined as zero when
$\|\mathbf p(X_n)\|_2\|\mathbf p(R_n)\|_2\leq10^{-12}$.
Thus, Flow Profile compares the temporal pattern of motion magnitude rather
than optical-flow direction or absolute visual similarity.

\paragraph{Action controllability.}
KCtrl uses constant-action counterfactual pairs from the same initial scene.
We sample frames $0,4,\ldots,76$, yielding $20$ sampled frames and $19$
consecutive relative poses. For action $a$, let
$v_{s,a,k}\in{\rm R}^{3}$ denote the translation component of relative pose
$k$ in scene $s$, where $k=0,\ldots,18$. Define the requested axis and sign as
$j(\mathrm{forward})=j(\mathrm{back})=z$,
$j(\mathrm{left})=j(\mathrm{right})=x$, and
$\eta_{\rm f}=\eta_{\rm r}=+1$,
$\eta_{\rm b}=\eta_{\rm l}=-1$. The net-direction indicator is
\begin{equation}
 C_{s,a}={\rm I}\!\left[
 \eta_a\sum_{k=0}^{18}(v_{s,a,k})_{j(a)}>0
 \right],
\end{equation}
where ${\rm I}[\cdot]$ denotes the indicator function.
With $S=77$ scenes, the main-table score averages the two opposite translation
pairs per scene:
\begin{equation}
 {\rm KCtrl}=\frac{1}{2S}\sum_{s=1}^{S}
 \left(C_{s,{\rm f}}C_{s,{\rm b}}+C_{s,{\rm l}}C_{s,{\rm r}}\right).
\end{equation}
Thus a pair receives credit only if both opposite commands yield the requested
net camera-motion sign; no test-time motion-magnitude threshold is used.

Mouse Accuracy follows the GameWorld/VPT inverse-dynamics evaluator
~\cite{zhang2025matrixgame}. Let $g_{\rm IDM}(X)_t\in{\rm R}^2$ be its
predicted camera action at frame $t$, and let
$q$ be GameWorld's nine-way camera-direction quantizer,
$q:{\rm R}^2\rightarrow\{0,\ldots,8\}$.
Write TL and TR for turn-left and turn-right,
respectively; their expected labels are $y_{\rm TL}=3$ and $y_{\rm TR}=4$. Over the two
turn actions, $S=77$ scenes, and $T_{\rm IDM}=76$ evaluated predictions per
clip, define the per-clip accuracy as
$\hat y_{s,a,t}=q(g_{\rm IDM}(X_{s,a})_t)$ and
\begin{equation}
 A_{s,a}=\frac{1}{T_{\rm IDM}}\sum_{t=0}^{T_{\rm IDM}-1}
 {\rm I}[\hat y_{s,a,t}=y_a].
\end{equation}
The reported score is
\begin{equation}
 {\rm MouseAcc}=\frac{1}{2S}\sum_{s=1}^{S}
 \left(A_{s,{\rm TL}}+A_{s,{\rm TR}}\right).
\end{equation}
This is a nine-way direction-classification accuracy under the commanded
constant mouse action, not a continuous regression error or a pixel-reference
metric.

\paragraph{Efficiency.}
For measured chunk $j$, let $L_j$ be synchronized sampling time in
milliseconds and $C_j$ the number of newly generated frames (12 for ForgeWM
and Matrix-Game 2.0, and 16 for HY-WorldPlay).  Over the $J$ timed chunks,
\begin{equation}
 {\rm Latency}=\frac{1}{J}\sum_{j=1}^{J}L_j,\qquad
 {\rm FPS}=\frac{1}{J}\sum_{j=1}^{J}\frac{1000C_j}{L_j}.
\end{equation}
Loading, VAE decoding, and file writing are excluded.  Efficiency columns are
reported for context and are not bolded in the table.

\paragraph{Profiling and resampling.}
Efficiency is profiled on NVIDIA GPUs over 30 clips after three
warm-ups; the four-step ForgeWM/Matrix-Game comparison pools 90 measurements
from three GPUs, and efficiency reporting follows the same 30-clip protocol.
No-reference metrics use scene-level bootstrap and paired metrics use
trajectory-level bootstrap; plots report means with bootstrap 95\% confidence
intervals.  The fixed-checkpoint step-scaling study freezes ForgeWM-1 and
evaluates $1/2/4/8/16/32$ steps on identical scene--action--seed tuples.
Replay-refinement evaluation starts from a saved one-step draft and refines it
with the same frozen student under the recorded controls ($r=0.3$, four-step
schedule), comparing against the draft and against from-noise controls.

\subsection{User Study Protocol}
\label{app:user-study}

We recruited $41$ student volunteers to participate in the study. Each was shown a sequence of comparisons; in
every comparison three clips appeared side by side, one from the four-step
ForgeWM student, one from Matrix-Game 2.0, and one from HY-WorldPlay.  All clips
in a comparison share the same source initial state and recorded control trace;
each model consumes that trace through its native or converted control interface
described above.  Model identities were hidden and
the left-to-right placement was randomized independently for each comparison,
so position carried no information about which system produced a clip.

Each participant judged five comparisons per criterion under three criteria,
giving $41\times5=205$ selections per criterion and $615$ in total.  The
criteria were described to participants as follows.  \emph{Visual Quality}:
sharpness, naturalness, level of detail, and absence of artifacts.
\emph{Action Accuracy}: whether the motion in the clip follows the direction and
the persistence of the control input.  \emph{Spatiotemporal Consistency}:
whether scene identity and geometry are preserved over time, and whether the
clip is free of flicker and warping.  For each comparison and criterion the
participant selected exactly one of the three clips; there was no tie option,
which is why the three percentages for a criterion sum to $100\%$.

\section{CrossFPS Evaluation and Adaptation}
\label{app:crossfps}

\subsection{Cross-Domain FPS Evaluation}
\label{app:crossfps-eval}

We sample 25 clips from each of the seven games in the official CrossFPS
evaluation split~\cite{tong2026scope} with a fixed seed, giving 175 clips.  Equal
counts per game matter here: the released split is heavily unbalanced (its largest
title carries roughly $65\times$ the clips of its smallest), so an unweighted
mean over clips would report the largest title's score under a cross-game label.
We therefore both sample equally and report a macro-average over games.  Since
the released archive does not separately label its validation and test
partitions, the sample is drawn from the combined evaluation split and these
values are not directly comparable to the benchmark's partition-specific
published results.

Each clip is replayed under its own recorded gamepad trace.  Generation and
scoring follow the same protocol as the Minecraft experiments: 81 pixel frames
at $640\!\times\!352$ from the clip's first frame, and paired LPIPS
(AlexNet backbone) on 16 frames sampled across the rollout, excluding the given
first frame. PSNR is computed and averaged over the same temporally aligned
frame pairs. Reference footage is temporally resampled to the same frame count.
Motion is measured as the mean dense optical-flow magnitude, the same estimator
as the main table's flow column; \emph{flow ratio} is the generated magnitude
divided by the reference magnitude, so $1.0$ means the rollout moves as much as
the reference.

\begin{table}[t]
    \centering
    {
    \small
    \setlength{\tabcolsep}{3.5pt}
    \begin{tabular}{lcccc}
        \toprule
        Game & LPIPS$\downarrow$ & PSNR$\uparrow$ & Flow & Ratio \\
        \midrule
        Xonotic                & 0.5828 & 11.21 & 13.88 & 1.66 \\
        Modern Warfare III     & 0.6352 & 11.61 &  8.15 & 1.35 \\
        Modern Warfare         & 0.6479 & 10.99 &  8.29 & 1.46 \\
        Warzone                & 0.6695 & 10.66 &  9.13 & 1.78 \\
        Halo Infinite          & 0.6730 &  9.75 & 11.23 & 1.46 \\
        Halo                   & 0.6920 &  9.31 & 13.28 & 1.29 \\
        Call of Duty           & 0.6933 &  9.67 & 10.40 & 1.16 \\
        \midrule
        Macro-average          & 0.6562 & 10.46 & 10.62 & 1.45 \\
        \bottomrule
    \end{tabular}
    }
    \caption{\textbf{Per-game CrossFPS results,} 25 clips per game.  Flow is
    the mean dense optical-flow magnitude in pixels; ratio is generated over
    reference, where $1.0$ matches the reference motion scale.}
    \label{tab:crossfps-pergame}
\end{table}

Table~\ref{tab:crossfps-pergame} gives the breakdown.  LPIPS spans
$0.583$--$0.693$ across titles.  Flow ratios range from $1.16$ to $1.78$, with a
macro-average of $1.45$, indicating systematic over-response relative to the
reference motion scale.

\subsection{CrossFPS Adaptation: Action Module and Training}
\label{app:crossfps-adaptation}

The CrossFPS checkpoint reported in the main paper follows the same four-stage
training recipe and retains the same backbone and action-module topology.
Adapting to gamepad control requires only an interface-level change: the
continuous-control input width is increased from two to four channels, while
the data and control semantics differ from the Minecraft setting. This section
specifies that interface adaptation and the initialization of the added input
channels.

\paragraph{Action schema.}
Minecraft control uses a two-dimensional mouse delta and a six-state keyboard
vector.  CrossFPS is driven by a gamepad, so the continuous channel widens from
two to four dimensions -- the $(x,y)$ deflection of the left analog stick
(movement) and the $(x,y)$ deflection of the right analog stick (camera look) --
while the discrete channel stays six-dimensional and is reinterpreted as the six
gamepad buttons (right trigger / fire, left trigger / aim-down-sights, and the
south / west / north / right-thumb buttons for jump, reload, weapon switch, and
melee).  In the model configuration, this changes only
\texttt{mouse\_dim\_in} from $2$ to $4$. The backbone, action-module topology,
and all downstream dimensions remain unchanged; only the width of the first
continuous-control input projection is enlarged.

\paragraph{Where the change lands.}
Continuous controls enter through the first layer of the action module's mouse
branch, a linear layer over the windowed, temporally-grouped control vector
concatenated with the visual hidden state (the \texttt{mouse\_mlp} input
projection).  Widening the continuous input from two to
four channels enlarges only this input projection -- from
$1536 + 2\!\times\!4\!\times\!3 = 1560$ to $1536 + 4\!\times\!4\!\times\!3 =
1584$ input units ($1536$ visual hidden units plus
\texttt{mouse\_dim\_in}\,$\times$\,VAE temporal compression\,$\times$\,window
control units) -- leaving every downstream weight shape unchanged.

\paragraph{Initialization of the new channels.}
A naive warm-start fails in two ways that we observed directly. First, the
shape filter used for checkpoint loading discards the \emph{entire} input
projection when its input width changes, including the $1536$ columns that read
the visual hidden state and the columns for the two original control channels;
a model re-initialized this way collapses to a ``use the hidden state, ignore
the control'' solution and does not recover action controllability during
training. Second, keeping the visual columns but \emph{zero-initializing} the
two new control channels leaves those channels without gradient signal at the
small Stage~0 learning rate: with no input variation reaching a zeroed column,
the symmetry is never broken and the two added channels stay effectively dead.

We therefore \emph{graft} the input projection from the base checkpoint. The
$1536$ visual columns and the two original control channels are copied
verbatim, and each of the two \emph{new} channels is initialized by copying one
of the pretrained control channels rather than from zero, so every channel
begins from a trained-scale, symmetry-broken state that gradient descent can
move. The zero-action-residual-at-initialization property is preserved not by
the input projection but by the action module's \emph{output} projection: the
base checkpoint ships the mouse and keyboard output projections zeroed, so at
the first update the widened action branch still contributes no residual and
the base video prior is left intact while the new interface is learned.

\paragraph{Training.}
Stage~0 adapts the grafted generator on the merged CrossFPS corpus
(65{,}246 sharded clips across the seven titles) for $12{,}000$ updates.
Stage~1 initializes from this Stage~0 checkpoint, which already contains
the four-channel continuous-control projection. Stage~2 then initializes
from the resulting Stage~1 checkpoint, and Stage~3 initializes from
Stage~2; no further grafting is required. Stages~1--3 otherwise follow
the corresponding Minecraft objectives and optimization settings in
Table~\ref{tab:training-hparams}
($20{,}000$ teacher-forced updates, $6{,}000$ consistency-distillation
updates, and $4{,}000$ Stage~3 trainer iterations). The four-step
CrossFPS student uses the $[1000,750,500,250]$ schedule and is evaluated
in Table~\ref{tab:crossfps-pergame}.

\section{Limitations}
\label{app:limitations}

We note three limitations that the results in this paper make visible.
\emph{Long-horizon drift.} The quantitative protocol evaluates a
$77$-frame window; the extended rollouts of
Figure~\ref{fig:long-rollouts}, roughly $3.5\times$ that horizon, show a
slow loss of block structure and spreading color artifacts at later timestamps.
Causalization and distillation reduce but do not eliminate the autoregressive
accumulation of error, and we do not claim indefinite rollout stability.
\emph{Motion over-response.} On the cross-domain FPS split the generated flow
magnitude exceeds the reference by a macro-average ratio of $1.45$
(Table~\ref{tab:crossfps-pergame}): the model tends to move more than the
recorded control implies, so magnitude fidelity lags directional and persistence
fidelity. \emph{Budget-dependent stage value.}
At the four-step budget, Stage~3 increases Imaging Quality but does
not improve paired LPIPS relative to Stage~2. Its measured incremental
effect at this budget is therefore metric-dependent; the stage-wise
ablation does not isolate its incremental effect at one or two steps. 

\end{document}